\documentclass[runningheads]{llncs}

\usepackage{eccv}

\usepackage{eccvabbrv}

\usepackage[accsupp]{axessibility}

\usepackage{amsmath}
\usepackage{amssymb}
\usepackage{bm}
\usepackage{graphicx}
\usepackage{booktabs}
\usepackage{multirow}
\usepackage{makecell}
\usepackage{array}
\usepackage{pifont}
\usepackage{tabularx}
\usepackage{subcaption}
\usepackage[table]{xcolor}
\usepackage{url}
\usepackage[utf8]{inputenc}
\usepackage{microtype}
\usepackage{tcolorbox}
\tcbuselibrary{breakable}
\usepackage{listings}

\usepackage{booktabs}   
\usepackage{graphicx}  
\usepackage{pifont}     
\usepackage{amssymb}
\newcommand{\cmark}{{\color{green!60!black}\ding{51}}} 
\newcommand{\xmark}{{\color{red!70!black}\ding{55}}} 
\usepackage{hyperref}
\usepackage{colortbl}
\usepackage{xcolor}
\usepackage{orcidlink}
\definecolor{skyblue}{RGB}{169, 217, 234}

\DeclareRobustCommand{\equalcontribmark}{\texorpdfstring{\textsuperscript{*}}{}} 
\DeclareRobustCommand{\corrauthormark}{\texorpdfstring{\textsuperscript{\ensuremath{\dagger}}}{}}

\makeatletter
\newcommand{\setauthorfootnotes}{%
  \renewcommand{\thefootnote}{%
    \ifcase\value{footnote}%
    \or *%
    \or \ensuremath{\dagger}%
    \or \ensuremath{\ddagger}%
    \else \arabic{footnote}%
    \fi}}
\makeatother

\setauthorfootnotes

\title{SciIR: A Large-scale Training Dataset and Benchmark for Scientific Image Reasoning Generation}

\titlerunning{SciIR}

\newcolumntype{C}{>{\centering\arraybackslash}X}

\author{Zhiyuan Ma\inst{1}\orcidlink{0009-0006-3756-5621}\equalcontribmark \and 
Zhengfeng Shi\inst{1,2}\equalcontribmark \and 
Yuning An\inst{1} \and 
Peize Li\inst{1} \and 
Jiabao Wei\inst{1} \and 
Ruijie Li\inst{1} \and 
Junhao Xiao\inst{1} \and 
Jianjun Li\inst{1}\orcidlink{0000-0002-5265-7624}\corrauthormark \and 
Bowen Zhou\inst{3}}
\authorrunning{Z.~Ma et al.}

\institute{School of Computer Science and Technology, Huazhong University of Science and Technology, China
\and
School of Airspace Science and Engineering, Shandong University, China
\and
Department of Electronic Engineering, Tsinghua University, China\\
\email{\{mzyth,jianjunli\}@hust.edu.cn, zhengfengshi@mail.sdu.edu.cn}}

\begin{document}
\maketitle
\vspace{-0.8em}
\begin{center}
{\normalsize
\href{https://github.com/MAIR-Lab-HUST/SciIR}
{\raisebox{-0.15em}{\includegraphics[height=0.9em]{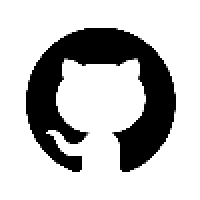}}\ \texttt{https://github.com/MAIR-Lab-HUST/SciIR}}
\\[0.35em]
\href{https://huggingface.co/datasets/MAIR-Lab-HUST/SciIR-82k}
{\raisebox{-0.15em}{\includegraphics[height=0.9em]{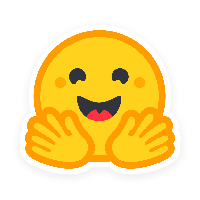}}\ \texttt{https://huggingface.co/datasets/MAIR-Lab-HUST/SciIR-82k}}
}
\end{center}
\vspace{0.5em}

\setauthorfootnotes
\footnotetext[1]{%
Equal contribution.}
\footnotetext[2]{%
Corresponding author.}

\captionsetup{hypcap=false}
\begin{center}
    \includegraphics[width=\textwidth]{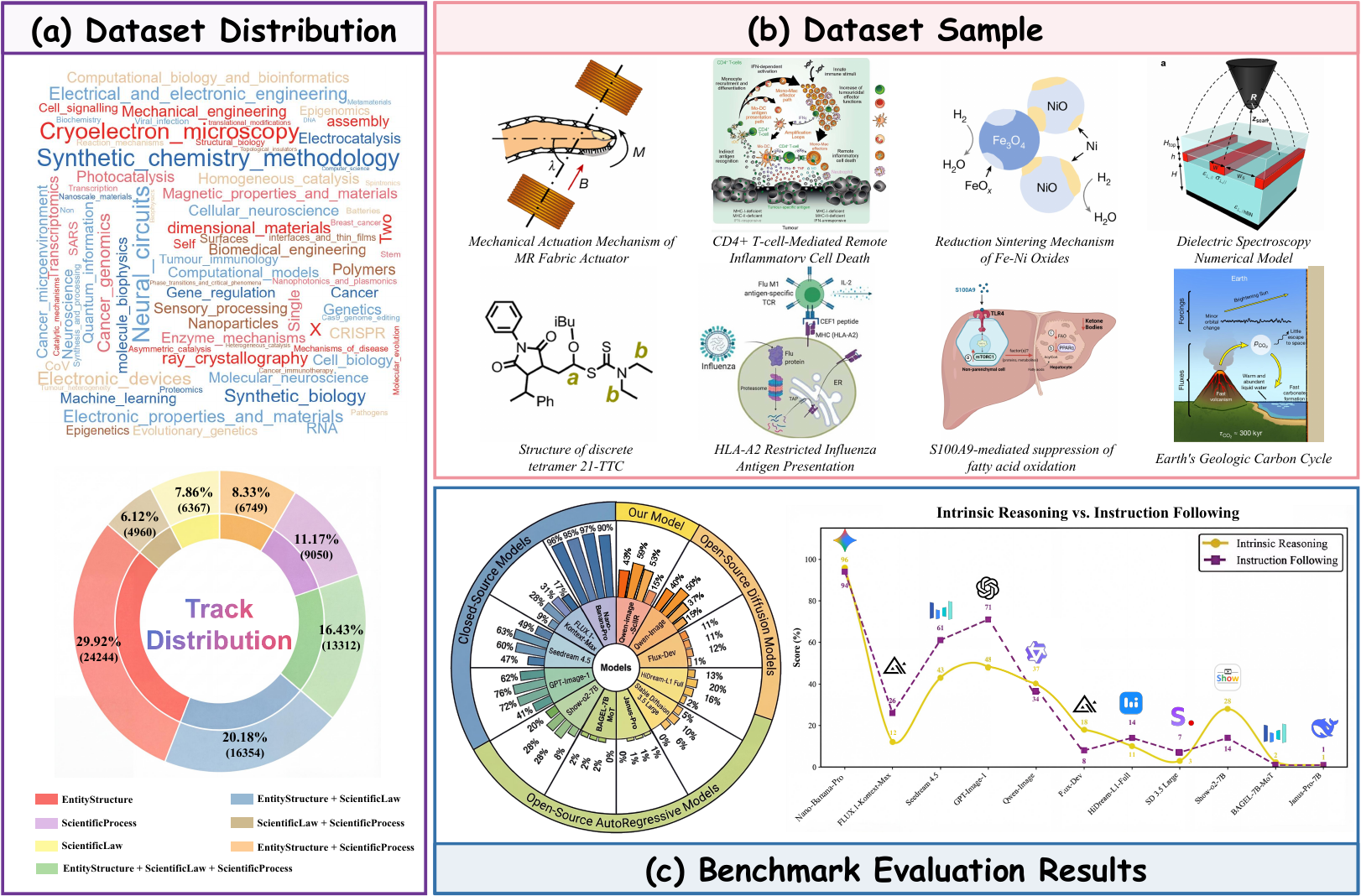}
    \captionof{figure}{\textbf{Overview of SciIR.} (a) SciIR-82k: keyword word cloud and distribution across semiotic-oriented image generation tracks. (b) Example figures from diverse domains. (c) Illustration of SciIR-Bench results across various open- and closed-source models with a comparison of Intrinsic Reasoning vs. Instruction Following.}
    \label{fig:overview}
\end{center}

\begin{abstract}

While Text-to-Image (T2I) models have shown remarkable success in generating photorealistic visual content, they still struggle with the rigorous semantic alignment and logical reasoning required for scientific imagery.
Inspired by Peirce's Semiotic Triad, we introduce Scientific Image Reasoning (SciIR), a comprehensive resource for training and evaluation of scientific image generation.
We formalize scientific reasoning into three core dimensions: Entity Structure (\emph{Icon}), Scientific Process (\emph{Index}), and Scientific Law (\emph{Symbol}).
Specifically, to overcome the scarcity of training data in scientific image generation, we elaborately create SciIR-82k, a large-scale dataset containing over 80,000 high-quality scientific image-text pairs from cutting-edge publications.
The dataset is hierarchically organized according to the semiotic dimensions and incorporates a Scientific Reasoning Chain-of-Thought (Sci-RCoT) to explicitly model underlying visual logic.
For evaluation, we propose SciIR-Bench, which aligns with these three semiotic levels and employs an Atomic Checklist to convert the outcome-oriented scientific accuracy into process-oriented, verifiable, fine-grained questions.
Our extensive experiments reveal significant deficiencies in current models' scientific reasoning capabilities.
Furthermore, by fine-tuning on the SciIR-82k dataset, we developed the Qwen-Image-SciIR model, which achieves a substantial improvement on the SciIR-Bench, increasing the final score from 35\% to 43\%, laying a solid foundation for future advances in scientific image generation.

\end{abstract}

\section{Introduction}
Recent advances in Text-to-Image (T2I) generation have yielded high-quality models: diffusion-based approaches \cite{yang2023improving,esser2024scaling,gao2024lumina,xie2025sana,qin2025lumina,ma2025efficient,ma2024neural} deliver exceptional visual realism and stylistic diversity, autoregressive methods \cite{sun2024autoregressive,zhang2024var,wang2024emu3,chen2025janus} excel at semantic alignment and complex instruction following, and emerging reasoning-augmented techniques integrate chain-of-thought (CoT) strategies \cite{liao2025imagegen,pan2025self} to help resolve ambiguities and map abstract concepts to concrete visual attributes. Despite these gains, T2I models remain fundamentally constrained when required to perform rigorous reasoning under complex, multi-constraint scenarios—most notably in scientific imagery, which must strictly adhere to physical laws, accurate topology, and causal logic. Even state-of-the-art closed-source systems (\eg, \texttt{Nano-Banana}) that achieve high perceptual fidelity and object-level consistency frequently violate domain-specific logical constraints, producing results that are ``visually plausible but factually incorrect''.

By contrast, open-source alternatives \cite{esser2024scaling,xie2024show,deng2025emerging,labs2025flux,wu2025qwen,cai2025hidream,ma2025janusflow} face more fundamental challenges: in knowledge-intensive contexts, they often struggle to internalize domain-specific knowledge and satisfy strict scientific constraints.
This divergence underscores a vital principle: \textit{Perceptual fidelity does not equate to reasoning robustness, and visual polish cannot compensate for the absence of scientific validity.}
These methodological gaps crystallize into three bottlenecks that impede scientific image generation:

\begin{enumerate}
  \item \textit{Data aspect — scarcity of logic-annotated resources.} High-quality scientific figures with explicit reasoning annotations are scarce because producing and annotating them requires deep, domain-specific expertise. Existing datasets therefore lack the ``visual logic'' necessary for models to learn the dependencies required to map text into scientifically accurate structures.
  \item \textit{Evaluation aspect — lack of scientific-correctness standards.} Current evaluation frameworks leave a structural gap in assessing scientific correctness. Although recent tools such as AutoFigure~\cite{zhu2026autofigure} and PaperBanana~\cite{zhu2026paperbanana} facilitate automated figure generation, their evaluation emphasizes layout and workflow rather than underlying scientific logic. Moreover, benchmarks (e.g., SridBench~\cite{chang2025sridbench}) often do not offer fine-grained semantic diagnosis, making it hard to define a verifiable ground truth for auditing the multidimensional constraints in scientific imagery.
  \item \textit{Model aspect — deficiencies in enforcing scientific constraints.} Open-source models in particular lack specialized scientific reasoning and therefore struggle to satisfy hard constraints such as topology or reaction causality, frequently producing hallucinations that violate basic physical laws.
\end{enumerate}

Motivated by the above three aspects, we propose \textbf{SciIR}, a semiotic triad-based dataset and benchmark designed to promote scientific rigor in image reasoning generation, as shown in \cref{fig:overview}.
In short, our main contributions are summarized as follows:
\begin{itemize}
  \item We construct \textbf{SciIR-82k}, a large-scale refined dataset of over $80,000$ scientific image-text pairs from \textit{Nature} and \textit{Nature Communications}. This dataset is enhanced with Sci-RCoT annotations, which explicitly formalize latent visual reasoning pathways to train models on underlying scientific logic.
  \item We introduce \textbf{SciIR-Bench}, the first benchmark to systematically categorize evaluation tracks based on multidimensional scientific correctness, employing a novel atomic checklist to provide fine-grained, verifiable questions.
  \item We develop \textbf{Qwen-Image-SciIR}, a strong open-source baseline that boosts the final score on SciIR-Bench from $35\%$ to $43\%$ by fine-tuning Qwen-Image-2512 on SciIR-82k, and serves as a reliable starting point for future research on scientific image reasoning generation.
\end{itemize}

\section{Related Work}
\begin{table}[t]
\caption{Comparison of SciIR-82k with representative T2I datasets.}
\label{tab:t2i_dataset_comparison}
\centering
\small
\setlength{\tabcolsep}{3pt}
\begin{tabular}{lccc}
\toprule
\textbf{Dataset} & \textbf{Scale} & \textbf{Text Length} & \textbf{Reasoning Type} \\
\hline
\rowcolor{gray!15} \multicolumn{4}{c}{\textit{\textbf{Synthetic Image Datasets}}} \\
JourneyDB\cite{sun2023journeydb} & 4M & Short & None \\
PixelProse\cite{pixelprose24} & 16M & Long & None \\
FLUX-Reason-6M\cite{fang2025flux} & 6M & Short & Visual \\
Science-T2I\cite{li2025science} & 20K & Short & Outcome-oriented \\
\midrule
\rowcolor{gray!15} \multicolumn{4}{c}{\textit{\textbf{Non-Synthetic Image Datasets}}} \\
CC-12M\cite{changpinyo2021cc12m} & 12M & Short & None \\
LAION-Aesthetics\cite{schuhmann2022laion} & 120M & Short & None \\
TextCaps\cite{sidorov2020textcaps} & 28K & Short & None \\
DOCCI\cite{OnoeDocci2024} & 15K & Long & None \\
\textbf{SciIR-82k (Ours)} & \textbf{82K} & \textbf{Short + Long} & \textbf{ Process-oriented} \\
\bottomrule
\end{tabular}
\end{table}

\subsection{Text-to-Image Datasets}

Current text-to-image (T2I) datasets generally lack the cognitive depth required for complex scientific synthesis. As shown in \cref{tab:t2i_dataset_comparison}, both synthetic and non-synthetic collections provide predominantly descriptive captions, omitting explicit reasoning chains or structured semantic relations. Specifically, while synthetic datasets \cite{sun2023journeydb,pixelprose24,fang2025flux,li2025science} provide large-scale, controllable supervision, they often inherit biases from source models—prioritizing visual plausibility and stylistic diversity over rigorous logical consistency. Conversely, non-synthetic datasets \cite{sidorov2020textcaps,changpinyo2021cc12m,OnoeDocci2024,schuhmann2022laion} source web image-text pairs to cover everyday concepts, but their typically short captions lack domain knowledge and explicit logical structures.
While scientific datasets like Science-T2I \cite{li2025science} leverage specialized knowledge to mitigate inconsistencies in generated diagrams, such early efforts remain limited in scale and prioritize final visual correctness. Consequently, their reliance on post-hoc preference modeling for implicit, outcome-oriented reasoning fails to support the intrinsic, process-oriented reasoning required during generation.
To bridge this gap, we introduce SciIR-82k. Sourced from academic publications, it is enriched with detailed Sci-RCoT annotations that formalize the logical steps for figure construction. By explicitly addressing complex scientific logic spanning Structure, Process, and Law—capturing structural relationships, causal mechanisms, and scientific principles—SciIR-82k provides process-oriented supervision. This explicit formalization allows models to learn rigorous, reasoning-conditioned visualizations from scratch, grasping not merely how scientific images look, but the underlying rationale behind their structured representations.

\begin{table}[tb]  
\caption{Comparison of SciIR-Bench with representative T2I benchmarks. \textbf{SL}: Scientific Law, \textbf{ES}: Entity Structure, \textbf{SP}: Scientific Process.}
\label{tab:sciir_comparison}
\centering
\scriptsize  
\setlength{\tabcolsep}{3pt} 
\renewcommand{\arraystretch}{1.1} 
\begin{tabular}{lllccccc} 
\toprule
\multirow{2}{*}{\textbf{Benchmark}} & \multirow{2}{*}{\textbf{Scale}} & \multirow{2}{*}{\textbf{Domain}} & \multicolumn{4}{c}{\textbf{Evaluation Dimensions}} & \multirow{2}{*}{\textbf{\shortstack{Fine-grained}}} 
\\ \cmidrule(lr){4-7}
& & & \makebox[0.9cm][c]{\textbf{SL}} & \makebox[0.9cm][c]{\textbf{ES}} & \makebox[0.9cm][c]{\textbf{SP}} & \makebox[0.9cm][c]{\textbf{Text}} & 
\\
\midrule
GenEval++\cite{ye2025echo}       & 280    & Generic      & \xmark & \cmark & \xmark & \xmark & \cmark \\
T2I-CompBench\cite{huang2023t2i}   & 6k  & Generic      & \xmark & \cmark & \xmark & \xmark & \cmark \\
WISE\cite{niu2025wise}             & 1k    & Generic      & \cmark & \xmark & \cmark & \xmark & \xmark \\
R2I-Bench\cite{chen2025r2i}        & 3068   & Generic      & \xmark & \xmark & \cmark & \xmark & \cmark \\
T2I-ReasonBench\cite{sun2025t2i} & 800  & Generic      & \cmark & \cmark & \xmark & \xmark & \cmark \\
ScImage\cite{zhang2024scimage}         & --   & Method. Diags.& \xmark & \cmark & \xmark & \cmark & \xmark \\
PaperBananaBench\cite{zhu2026paperbanana}& 292   & Method. Diags.& \xmark & \cmark & \cmark & \cmark & \xmark \\
FigureBench\cite{zhu2026autofigure}     & 3300   & Method. Diags.& \xmark & \cmark & \cmark & \cmark & \xmark \\
SridBench\cite{chang2025sridbench}       & 1120   & Sci. Illus.  & \cmark & \cmark & \xmark & \cmark & \xmark \\
SciGenBench\cite{lin2026scientific}     & --    & Sci. Illus.& \cmark & \cmark & \xmark & \cmark & \cmark \\
\midrule
\textbf{SciIR-Bench (Ours)} & \textbf{800} & \textbf{Sci. Illus.} & \textbf{\cmark} & \textbf{\cmark} & \textbf{\cmark} & \textbf{\cmark} & \textbf{\cmark} \\ 
\bottomrule
\end{tabular}
\end{table}

\subsection{Text-to-Image Benchmark}
Existing benchmarks evaluate generation through three evolving perspectives. \textit{1) Perceptual Quality:} Early metrics like IS \cite{salimans2016improved} and FID \cite{heusel2017gans} assess distributional realism. \textit{2) Prompt Alignment:} Benchmarks such as T2I-CompBench \cite{huang2023t2i} and GenEval++ \cite{ye2025echo} quantify textual fidelity via VLM-based scoring \cite{hessel2021clipscore, ghosh2023geneval, hu2023tifa}. \textit{3) Semantic Plausibility:} Recent frameworks (\eg, WISE \cite{niu2025wise}, T2I-ReasonBench \cite{sun2025t2i}) target logical reasoning using LLMs. Within the domain of scientific figure generation, PaperBananaBench \cite{zhu2026paperbanana} and FigureBench \cite{zhu2026autofigure} concentrate on the evaluation of flowcharts and statistical diagrams. While SridBench \cite{chang2025sridbench} specifically targets scientific diagrams, it remains focused on broad interpretation rather than fine-grained generative constraints.

To address this structural opacity, we posit that scientific correctness cannot be treated as a monolithic metric but requires systematic decomposition. Drawing inspiration from Peirce's Semiotic Triad \cite{peirce1931collected}, we decompose scientific correctness into \textit{law}, \textit{structure}, and \textit{process}. However, we observe that no existing benchmark comprehensively covers all three dimensions (as shown in \cref{tab:sciir_comparison}), failing to diagnose specific atomic violations (\eg, broken causal links). To bridge this gap, we propose SciIR-Bench, a diagnostic framework that holistically assesses these dimensions through explicit, verifiable criteria.

\subsection{Methods for Scientific Image Generation}
Recent research has also advanced the automated generation of scientific illustrations. For example, AutoFigure\cite{zhu2026autofigure} focuses on producing publication-ready illustrations from long-form text, emphasizing layout and aesthetics, while PaperBanana specializes in automating and improving the visualization quality of methodological workflows \cite{zhu2026paperbanana}. Alternatively, ImgCoder\cite{lin2026scientific} prioritizes programmatic synthesis (code generation) to circumvent pixel-level reasoning challenges, though it lacks the visual expressivity required for rendering nuanced or complex graphic details. Despite these advances, such efforts remain primarily confined to procedural and architectural diagrams. In contrast, our work targets scientific schematics that encode underlying natural principles (\eg, physical laws and causal mechanisms). Moving beyond layout fidelity, we explicitly model and evaluate \emph{scientific correctness}, requiring models not only to reproduce structural relationships but also to internalize and faithfully represent the domain-specific principles governing the depicted phenomena.

\section{SciIR-82k Dataset}
\begin{figure*}[t]
    \centering
    \includegraphics[width=\textwidth]{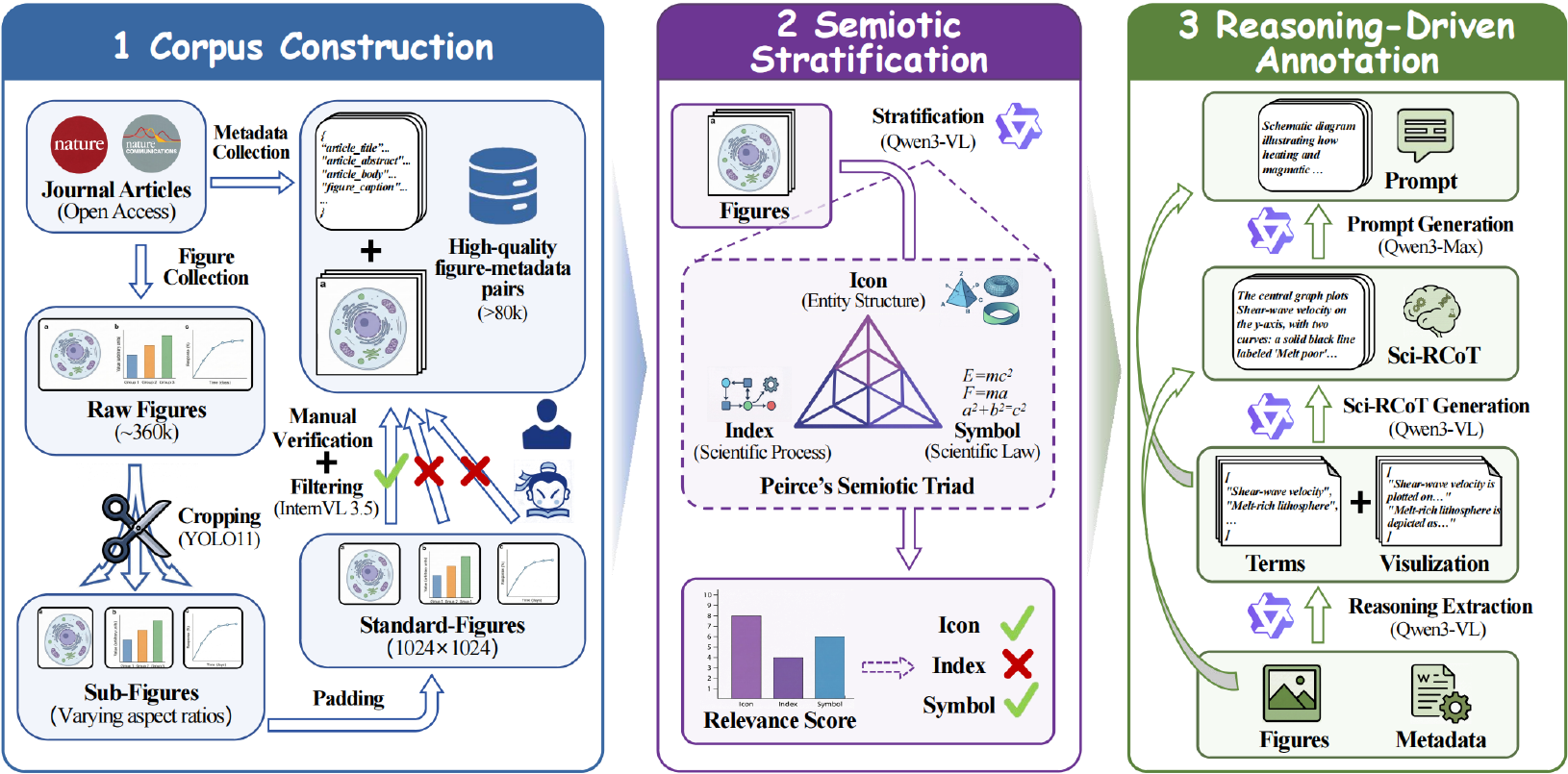}
\caption{\textbf{Overview of the SciIR-82k pipeline grounded in Peirce’s Semiotic Triad\cite{peirce1931collected}.} The framework comprises three stages: (1) Corpus Construction, employing YOLO11 and InternVL 3.5 for sub-figure extraction and filtering; (2) Semiotic Stratification, categorizing samples into Icon, Index, and Symbol tracks; and (3) Reasoning-Driven Annotation, which leverages Qwen3 models to reverse-engineer Sci-RCoT and prompts.}
    \label{fig:wide-example}
\end{figure*}

To systematically evaluate the scientific reasoning abilities of T2I models, we introduce SciIR-82k, a large-scale dataset of over $80,000$ high-quality scientific image-text pairs with complete annotations. As shown in \cref{fig:wide-example}, our SciIR-82k is grounded in a semiotic triad and constructed through a multi-stage automated pipeline for promoting scientific fidelity.  

\subsection{Theoretical Taxonomy: Semiotic Triad}
\label{sec:Semiotic Triad}

Scientific images are not merely visual imagery but abstract structures encoding logical relations and physical constraints.
To effectively formalize these, we ground our dataset in Peirce's Semiotic Triad~\cite{peirce1931collected}, \ie, \textit{Icon}, \textit{Index}, and \textit{Symbol}, which respectively correspond to the cognitive layers for scientific reasoning: \textit{1) Entity Structure} corresponds to \textit{iconic} representation via topological fidelity, which evaluates the geometric hierarchical reconstruction and spatial alignment of scientific entities.
\textit{2) Scientific Process} corresponds to \textit{indexical} representation indicating causal or temporal correlations, involving state transitions, experimental workflows, or causal chains.
\textit{3) Scientific Law} corresponds to \textit{symbolic} representation governed by abstract rules, ensuring adherence to fundamental laws (\eg, conservation of energy, molecular valence).
This taxonomy transcends simple visual-text alignment, establishing a structured framework for deep scientific reasoning.

\subsection{Corpus Construction}

Our corpus comprises articles licensed under CC BY 4.0 from \textit{Nature} and \textit{Nature Communications} to ensure authority; comprehensive compliance and provenance details are outlined in Appendix A.
From about $360$k raw figures, we employ Ultralytics-YOLO11 \cite{jocher2024ultralytics} as an automated layout analyzer to decompose multi-panel figures into semantically independent subfigures, which are then standardized to a $1024\times1024$ resolution.
Afterwards, a two-stage filtering pipeline---VLM-based screening with InternVL3.5 \cite{wang2025internvl3} to retain corresponding schematics, followed by manual verification---finally extracts over $80$k high-quality subfigures and their corresponding captions and content.
Data processing details are provided in the Appendix B.

\subsection{Semiotic Stratification}

We implement a stratification strategy to align the raw data with our semiotic taxonomy (\cref{sec:Semiotic Triad}).
Using Qwen3-VL \cite{Qwen3-VL} as a domain-specific evaluator, we assess each sample’s relevance score $s\in[1,10]$ to the three reasoning tracks (\texttt{Entity Structure}, \texttt{Scientific Process}, or \texttt{Scientific Law} ).
This categorization serves as a routing mechanism, directing images to targeted annotation pipelines according to their dominant semiotic attributes.

\subsection{Reasoning-Driven Annotation}
\label{sec:semiotic_pipeline}

In contemporary text-to-image and multimodal generation frameworks, the process typically follows a forward pipeline: Abstract Prompt $\rightarrow$ Logical Deduction $\rightarrow$ Visual Output, where high-level semantic intent is progressively instantiated into concrete visual representations.
For high-fidelity reasoning annotation, we invert this chain via \textit{logical reverse engineering}: reconstructing latent scientific reasoning (\ie, Sci-RCoT) from ground-truth images to derive precise prompts, with Qwen3-VL \cite{Qwen3-VL} for visual grounding and Qwen3-Max \cite{qwen3} for semantic abstraction.

\textbf{Taxonomy-Driven Reasoning Extraction.} In the first phase, Qwen3-VL \cite{Qwen3-VL} extracts taxonomy-guided structured information, prioritizing visual evidence (Image $>$ Caption $>$ Text).
The model parses input into JSON, decoupling: \textit{1) Terms:} Image-validated entities and nomenclatures. \textit{2) Visualization:} Descriptions of visual grounding (\eg, geometry, layout).
This taxonomy-driven extraction enforces semantic decomposition of scientific content, reduces hallucination, and converts free-form multimodal understanding into controllable, fine-grained reasoning units suitable for downstream transformation.

\textbf{Sci-RCoT Generation.} Afterwards, Qwen3-VL \cite{Qwen3-VL} synthesizes a Scientific Reasoning CoT (Sci-RCoT) by re-examining the image to integrate visual style (\eg, schematic diagrams) and text rendering requirements with ``Visualization'' entries.
 By transforming discrete reasoning units into a continuous visual reconstruction process, Sci-RCoT, as an explicit reasoning trace, bridges symbolic reasoning and holistic scene composition,  elucidating the causal logic underlying the mapping of abstract concepts to concrete spatial arrangements.

\textbf{Prompt Generation.} Eventually, Qwen3-Max \cite{qwen3} distills Sci-RCoT into a concise prompt via a ``Term-Substitution'' strategy: Visualization descriptions are replaced with canonical scientific terms while preserving the original visual style as the leading phrase, and only explicitly required textual renderings are retained. This produces a 
synchronized pair of abstract prompt and retained text. This step removes redundant visual detail while preserving scientific semantics, yielding a compact yet information-complete prompt representation. The resulting abstraction enhances controllability, improves semantic consistency across samples, and provides high-quality structured supervision for process reasoning-aware image generation and evaluation.

Overall, this pipeline performs logical reverse engineering from images to prompts by progressively transforming visual evidence into structured reasoning traces and finally into compact prompt representations. This design enforces explicit alignment between image structures, textual elements, and reasoning units, producing controllable and logically grounded prompts. The resulting annotations provide reliable supervision for evaluating and training process reasoning-aware scientific image generation models.

\section{SciIR-Bench}
\begin{figure*}[t]
    \centering
    \includegraphics[width=\textwidth]{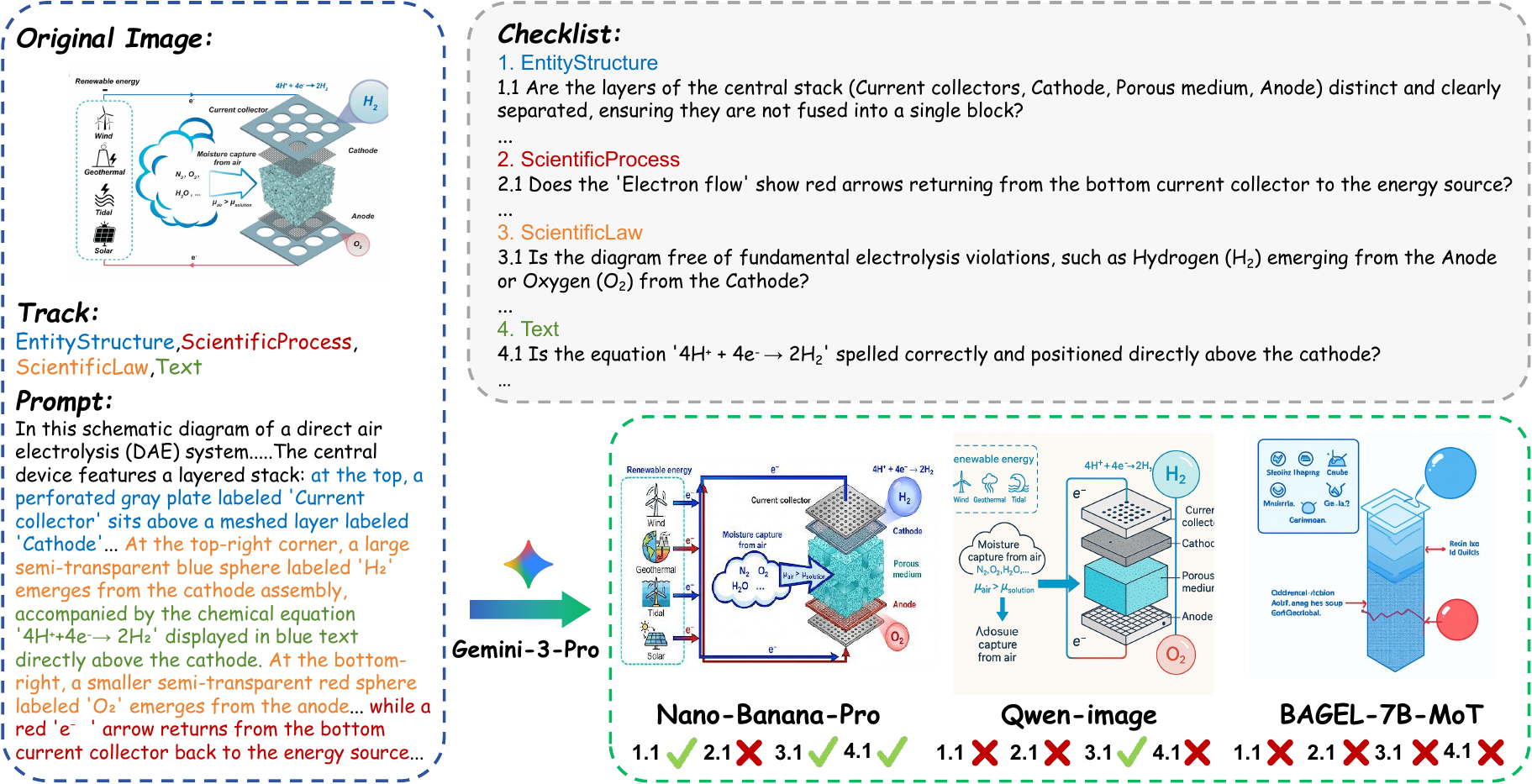}
    \caption{\textbf{An evaluation instance from SciIR-Bench.} Prompt from a sample covering all four tracks is used to guide various models in generating images. The output is then scrutinized by Gemini-3-Pro using a dimension-specific atomic checklist.}
    \label{fig:wide-example1}
\end{figure*}

To systematically evaluate the scientific reasoning and generation capabilities of current text-to-image models, we develop the SciIR-Bench (\cref{fig:wide-example1}). This benchmark moves beyond traditional holistic image quality metrics and instead measures whether models can faithfully instantiate structured scientific content—such as correctly rendering labeled entities, preserving spatial and topological relations, and accurately depicting multi-stage processes—without introducing unsupported elements or logical contradictions in the visualization.

\subsection{Evaluation Benchmark}

\textbf{Candidate Selection and Filtration.} From the massive corpus of processed samples, we distilled a high-quality evaluation benchmark consisting of 800 test instances. To ensure the benchmark challenges the upper limits of current models, we enforced a rigorous protocol focusing on both the breadth of scientific domains and the depth of reasoning complexity. We prioritized samples exhibiting High Term Density (term count $>3$) to guarantee sufficient semantic content. Details of the screening process are provided in the Appendix C. Furthermore, to verify multimodal reasoning capabilities, we selected candidates that necessitate compound reasoning across our theoretical dimensions, filtering out samples that do not contain valid reasoning paths in at least two of the three semiotic tracks (\texttt{Entity Structure}, \texttt{Scientific Process}, or \texttt{Scientific Law}).

\textbf{Taxonomy-Based Grouping.} To systematically evaluate model performance across different reasoning intersections, we exploit a four-folds strategy to categorize the filtered candidates into four distinct evaluation groups ($N=200$ per group). The first group represents the most complex ``holistic'' reasoning scenario, containing samples that simultaneously encompass attributes of all three tracks: Scientific Law, Entity Structure, and Scientific Process. The remaining three groups are constructed to test pairwise reasoning capabilities, covering the specific intersections of Law-Entity, Law-Process, and Entity-Process, respectively. This combinatorial approach ensures that the benchmark evaluates not just isolated knowledge but the model’s ability to synthesize conflicting constraints from multiple semiotic dimensions.

\textbf{Adaptive Difficulty Stratification.} Within each group, we bifurcated samples into two difficulty levels based on scientific term density to strictly disentangle \textit{Instruction Following} from \textit{intrinsic reasoning}:
\textit{1) Instruction Following (IF).}
Samples with high term density ($>$ median) are paired with the detailed Sci-RCoT. For these semantically saturated images, compressing the input into a concise prompt inevitably leads to \textit{semantic erosion}, causing the model to omit critical scientific details. By providing the exhaustive Sci-RCoT, we eliminate the ambiguity space, thereby strictly testing the model's fidelity in visualizing complex, fine-grained instructions.
\textit{2) Intrinsic Reasoning (IR).}
Conversely, samples with lower term density are paired with the abstract Prompt. In this regime, the input provides only high-level nomenclature without visual cues. This information sparsity compels the model to bridge the gap using latent domain knowledge, effectively evaluating its capacity to autonomously reason out valid spatial layouts and causal logic from abstract concepts.

\subsection{Evaluation Metrics}
\label{sec:metric}
Traditional generative metrics such as FID\cite{heusel2017gans} or CLIPScore\cite{radford2021learning} focus primarily on image fidelity or broad semantic similarity. However, these metrics fail to capture the factual correctness and logical consistency essential for scientific modeling. To bridge this gap, we propose a fine-grained, interpretable evaluation protocol designed to mimic the rigor of human peer review. Unlike black-box scoring, this VLM-driven checklist operationalizes evaluation through a transparent, three-stage automated pipeline comprising ground truth extraction, atomic questioning, and evidence-based refereeing.

\textbf{Atomic Checklist Generation.} We utilize the structured Reasoning content (extracted in Phase 1) as the absolute ground truth, avoiding reliance on potentially noisy reference images. To ensure comprehensive coverage, the checklist generation is strictly term-driven: for every ``Scientific Term'' identified in the reasoning structure, a VLM (\texttt{Gemini-3-Pro}) generates a corresponding binary validation query. This enforces Atomicity, as each question is strictly scoped to verify the visual manifestation of a single semantic unit (\eg, the specific morphology of a protein or the directionality of a process arrow). Moreover, to address the issue of hallucination, we generate supplementary adversarial questions tailored to each specific track to explicitly probe for domain-specific fabrications (\eg, non-existent chemical bonds), ensuring the model is penalized for inventing scientifically invalid details.

\textbf{Automated Evaluation.} In the adjudication phase, an advanced VLM acts as a ``Senior Scientific Reviewer'' to evaluate the generated images against the checklist. To prevent hallucinated judgments, we enforce a Visual Evidence Retrieval protocol: the referee must explicitly locate and describe the specific visual element mentioned in the query before assigning a verdict. The evaluation logic is strictly compartmentalized by category—Text is judged on spelling and positional exactness, while scientific tracks (\texttt{Entity Structure}, \texttt{Scientific Process}, \texttt{Scientific Law}) are evaluated on topological and causal logic. This separation ensures graphical errors result in scientific penalties, while textual errors are isolated to the text score.

\textbf{Accuracy Score.} Distinct from standard metrics that average accuracy across all questions, we adopt a rigorous sample-level pass rate to reflect the intolerance for error in scientific communication. Formally, let $Q_{i,c}$ be the set of atomic questions generated for a sample $i$ within a specific reasoning category $c$ (\eg, \texttt{Scientific Process}), and let $s(q) \in \{0,1\}$ denote the binary score of a single question $q$. The validity of a sample $i$ in category $c$, denoted as $V_{i,c}$, is defined by a strict veto mechanism:

\begin{equation}
V_{i,c} = \begin{cases} 
1 & \text{if } \prod_{q \in Q_{i,c}} s(q) = 1 \\
0 & \text{otherwise}
\end{cases}
\end{equation}

A sample is considered valid for a specific track only if it passes every atomic check associated with that track; a single failure renders the sample scientifically compromised for that dimension. The final accuracy score for category $c$ is calculated as the percentage of valid samples across the dataset $D$:

\begin{equation}
\text{Accuracy Score}_{c} = \frac{1}{|D|} \sum_{i \in D} V_{i,c}
\end{equation}

This strict metric provides a granular and uncompromising diagnostic of model performance, ensuring that high scores reflect true scientific robustness rather than partial hallucinatory success.
\begin{table}[t]
\caption{\textbf{Evaluation on SciIR-Bench.} We report the Accuracy Score (\%) for Intrinsic Reasoning (IR), Instruction Following (IF), and overall performance across four distinct tracks. \textbf{SL}: Scientific Law, \textbf{ES}: Entity Structure, \textbf{SP}: Scientific Process.}
\label{tab:main_results}
\centering
\small
\setlength{\tabcolsep}{3pt} 
\begin{tabularx}{\textwidth}{@{} l *{3}{C} @{\hspace{1.2em}} *{3}{C} @{\hspace{1.2em}} *{3}{C} @{\hspace{1.2em}} *{3}{C} c @{}}
\toprule
\multirow{2}{*}{\textbf{Model}} &
\multicolumn{3}{c}{\textbf{SL (\%)}} &
\multicolumn{3}{c}{\textbf{ES (\%)}} &
\multicolumn{3}{c}{\textbf{SP (\%)}} &
\multicolumn{3}{c}{\textbf{Text (\%)}} &
\multirow{2}{*}{\textbf{Final (\%)}} \\
\cmidrule(lr){2-4} \cmidrule(lr){5-7} \cmidrule(lr){8-10} \cmidrule(lr){11-13}
& IR & IF & Avg. & IR & IF & Avg. & IR & IF & Avg. & IR & IF & Avg. & \\
\midrule

\rowcolor{gray!15} \multicolumn{14}{c}{\textit{\textbf{Closed-Source Models}}} \\
Nano-Banana-Pro & 95 & 97 & 96 & 98 & 94 & 95 & 98 & 97 & 97 & 92 & 89 & 90 & 95 \\
FLUX.1-Kontext-Max & 16 & 19 & 17 & 15 & 40 & 31 & 13 & 36 & 28 & 3 & 11 & 9 & 22 \\
Seedream 4.5 & 39 & 57 & 49 & 56 & 67 & 63 & 53 & 64 & 60 & 25 & 55 & 47 & 55 \\
GPT-Image-1 & 52 & 72 & 62 & 64 & 82 & 76 & 51 & 82 & 72 & 23 & 47 & 41 & 62 \\

\midrule
\rowcolor{gray!15} \multicolumn{14}{c}{\textit{\textbf{Open-Source Diffusion Models}}} \\
Qwen-Image-2512 & 38 & 42 & 40 & 53 & 46 & 50 & 42 & 32 & 37 & 16 & 14 & 15 & 35 \\
Flux-Dev & 25 & 6 & 11 & 23 & 10 & 11 & 19 & 14 & 12 & 3 & 1 & 1 & 9 \\
HiDream-L1-Full & 12 & 14 & 13 & 14 & 23 & 20 & 15 & 16 & 16 & 1 & 2 & 2 & 13 \\
SD 3.5 Large & 3 & 7 & 5 & 6 & 12 & 10 & 2 & 8 & 6 & 0 & 0 & 0 & 5 \\

\midrule
\rowcolor{gray!15} \multicolumn{14}{c}{\textit{\textbf{Open-Source AutoRegressive Models}}} \\
Show-o2-7B & 28 & 12 & 20 & 42 & 15 & 28 & 32 & 25 & 28 & 12 & 4 & 8 & 21 \\
BAGEL-7B-MoT & 1 & 2 & 2 & 4 & 1 & 2 & 3 & 1 & 2 & 0 & 0 & 0 & 2 \\
Janus-Pro-7B & 1 & 2 & 1 & 1 & 2 & 1 & 1 & 1 & 1 & 0 & 0 & 0 & 1 \\

\midrule
\rowcolor{gray!15} \multicolumn{14}{c}{\textit{\textbf{Fine-tuned Models (Ours)}}} \\
\textbf{Qwen-Image-SciIR} & \textbf{37} & \textbf{50} & \textbf{43} & \textbf{56} & \textbf{62} & \textbf{59} & \textbf{52} & \textbf{54} & \textbf{53} & \textbf{14} & \textbf{15} & \textbf{15} & \textbf{43} \\

\bottomrule
\end{tabularx}
\end{table}
\section{Experiments and Analyses}

In this section, we provide a comprehensive analysis of model performance on SciIR-Bench. We discuss the quantitative results reported in \cref{tab:main_results}, analyze the correlation between our metrics and traditional evaluation standards, and qualitatively categorize common failure modes based on the proposed semiotic taxonomy.

\subsection{Experimental Settings}
\begin{figure*}[t]
    \centering
    \includegraphics[width=\textwidth]{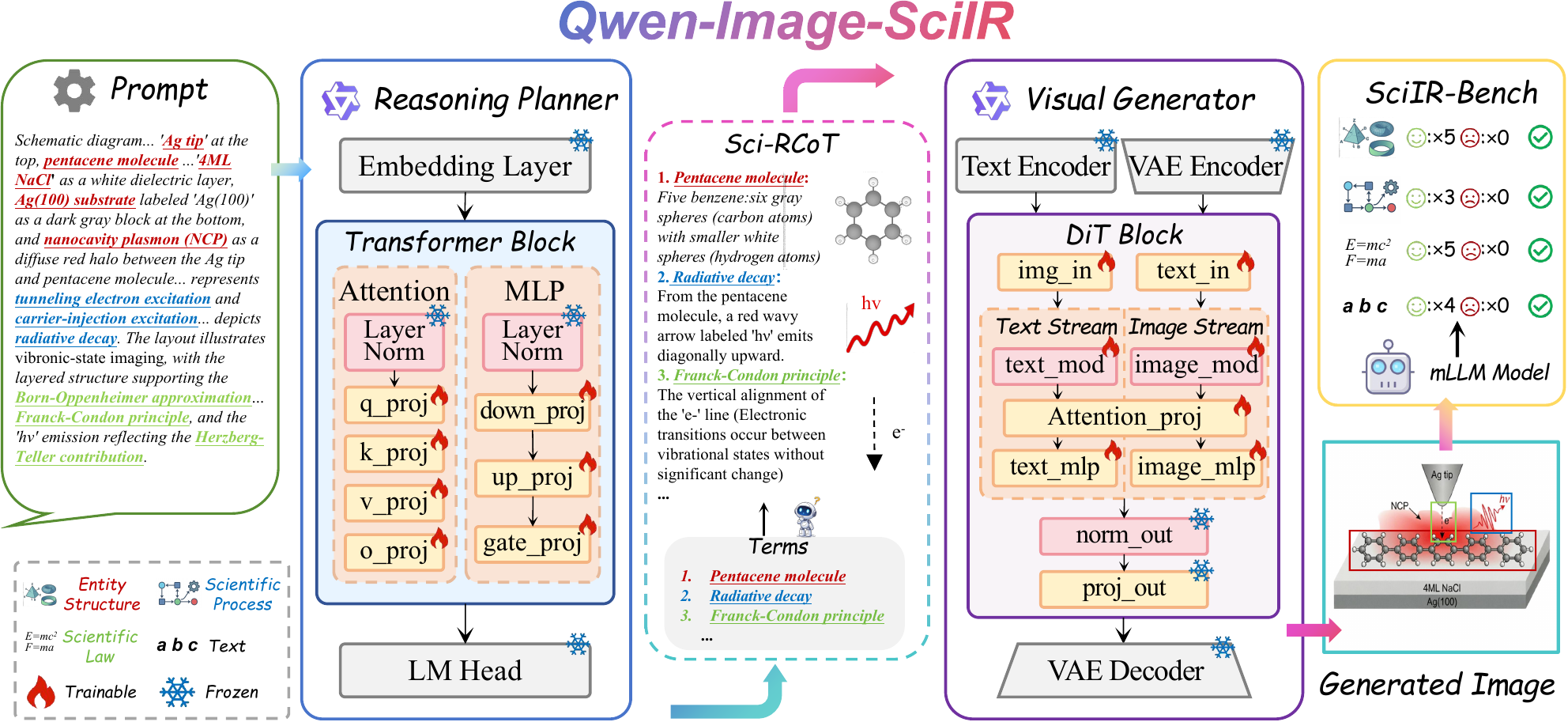}
    \caption{Qwen-Image-SciIR model architecture.}
    \label{fig:model}
\end{figure*}
\textbf{Implementation of Qwen-Image-SciIR.} Qwen-Image-SciIR is implemented to ensure rigorous zero-shot evaluation: we removed the 800 test instances in SciIR-Bench from the SciIR-82k corpus to avoid data leakage. As shown in Figure~\ref{fig:model}, the pipeline decouples scientific reasoning from visual synthesis via two fine-tuned modules. The first, Qwen2.5-7B-Instruct, serves as a reasoning planner and was fine-tuned on (prompt, Sci-RCoT) pairs using an all-linear LoRA configuration ($r=64, \alpha=16$). Specifically, LoRA adapters were integrated into all linear transformation layers within the Transformer blocks to maximize adaptation capacity. This module was trained with a learning rate of $1\times10^{-4}$ and a maximum context window of 2,048 tokens for one optimization step. The second, Qwen-Image-2512 as a visual generator, was fine-tuned on (Sci-RCoT, image) pairs via LoRA ($r=32$) applied to the diffusion transformer layers, with a learning rate of $1\times10^{-4}$, training resolution $1024\times1024$, and trained for one optimization step.

\textbf{Inference Protocol.} We develop a systematic inference pipeline for Qwen-Image-SciIR. Across 800 evaluation samples, encompassing both Intrinsic Reasoning (IR) and Instruction Following (IF) categories, we employ a chained generation flow. Specifically, the Reasoning Planner first infers a comprehensive Sci-RCoT from the input prompt, which is then utilized by the Visual Generator to synthesize the final image. This unified protocol ensures that the reasoning module is actively engaged for every instance, maintaining a standard reasoning-to-rendering process throughout the entire benchmark evaluation.

\subsection{Main Quantitative Results}
\cref{tab:main_results} presents the systematic evaluation of 12 T2I models across the SciIR-Bench. Our analysis reveals several key insights regarding the current landscape of scientific visual synthesis.

\textbf{Closed- \textit{vs.} Open-Source.}
Nano-Banana-pro's near-saturation performance (95\%) provides strong evidence that the task is solvable, yet a 60\% gap remains for open-source contenders. Critically, aesthetic-focused baselines like Flux-Dev fail (<10\%) on strict tracks, confirming a fundamental misalignment: current open-source training optimizes for \textit{perceptual fidelity}, sacrificing the logic essential for scientific accuracy.

\textbf{Instruction Following \textit{vs.} Intrinsic Reasoning.}
For the majority of models (\eg, GPT-Image-1, Seedream 4.5), performance under explicit Sci-RCoT prompting (IF) significantly outpaces abstract prompting (IR). For instance, FLUX.1-Kontext-Max's accuracy drops from 36\% to 13\% without dense guidance. This confirms that while they excel at executing detailed instructions, they lack internalized scientific world models to autonomously derive constraints. However, a counter-intuitive trend emerges in some open-weights models (\eg, Flux-Dev, Show-o2-7B), where IR outperforms IF. Rather than indicating superior reasoning, this highlights their deficiency in complex instruction adherence. Dense Sci-RCoT prompts overwhelm them, causing prompt overflow and attribute confusion. Thus, they paradoxically perform better on shorter abstract prompts by relying on superficial parametric memory.

\textbf{AutoRegressive \textit{vs.} Diffusion.}
Diffusion models currently maintain a distinct advantage over AutoRegressive (AR) architectures, with Qwen-Image-2512 (35\%) establishing a clear 14\% performance gap over the leading AR contender, Show-o2-7B (21\%).However, despite this architectural disparity, both paradigms share a critical vulnerability: severe failure in the Text track. With even the top models scoring merely 15\% (Diffusion) and 8\% (AR) on text generation, the results confirm a shared fundamental limitation: whether utilizing continuous denoising or discrete next-token prediction, current open-source visual generation frameworks fundamentally lack the fine-grained typographic control necessary for accurate scientific illustrating. 

\begin{figure*}[t]
    \centering
    \includegraphics[width=\textwidth]{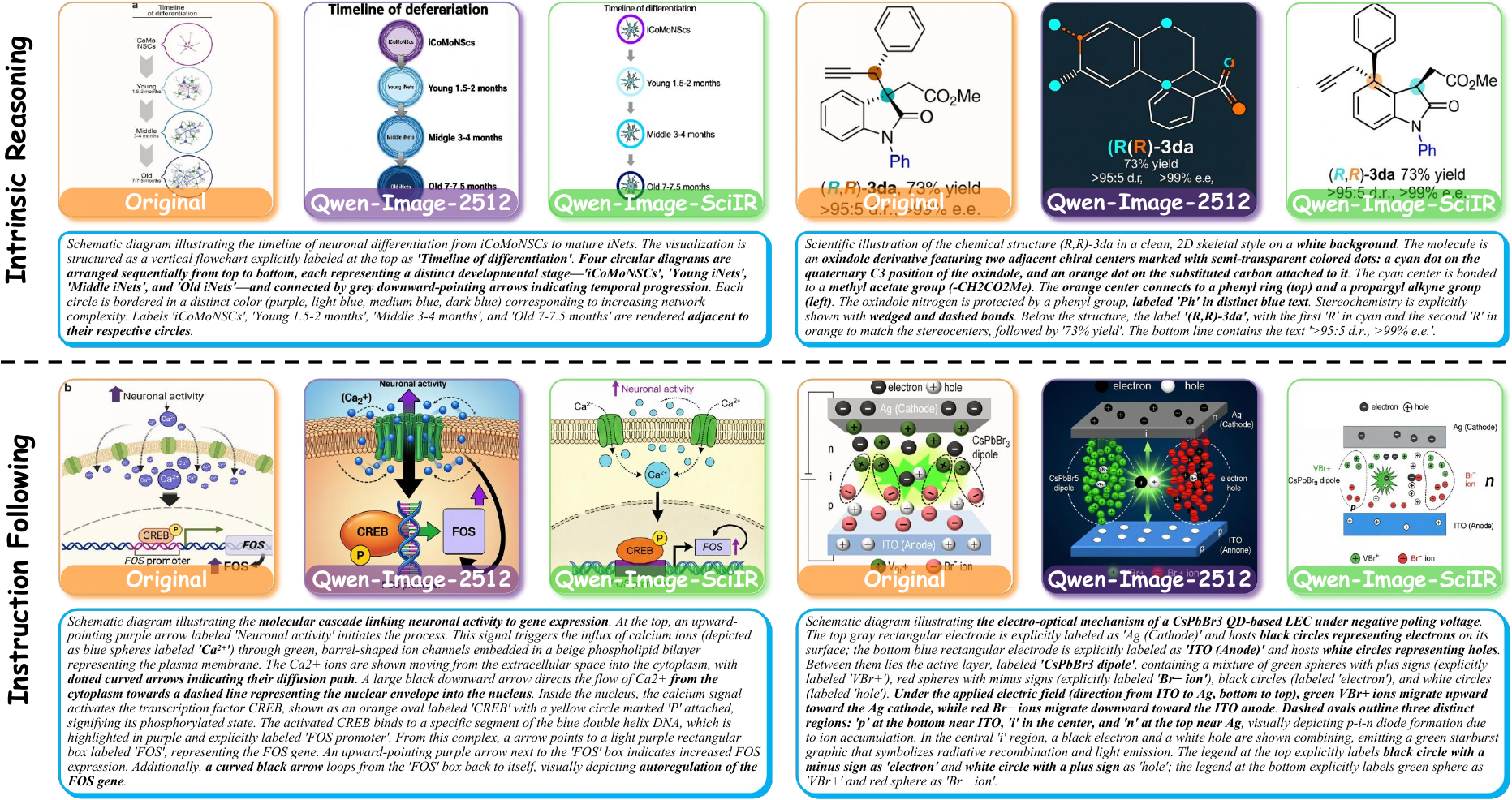}
    \caption{Qualitative comparison of generated results.}
    \label{fig:qualitative_comparison}
\end{figure*}

\textbf{Efficacy of Fine-tuning.} To validate the effectiveness of our proposed pipeline, we conduct a direct comparison between the fine-tuned Qwen-Image-SciIR and its backbone, Qwen-Image-2512. Quantitatively, \cref{tab:main_results} demonstrates a substantial performance gain, elevating the Final Score from 35\% to 43\%. This improvement is particularly pronounced in the \texttt{Scientific Process} and \texttt{Entity Structure} tracks, with increases of 16\% and 9\% respectively, indicating a robust enhancement in modeling sequential process and topological integrity.

\subsection{Qualitative Comparison}
Qualitatively, visual comparisons in \cref{fig:qualitative_comparison} reveal that Qwen-Image-SciIR shifts from a generic artistic style to a precise scientific illustration standard. We observe that the baseline Qwen-Image-2512 is prone to scientific hallucinations across three dimensions. 
The first is the \textbf{inability to correctly depict scientific processes}: it fails to visually depict the morphological progression of neuron development (top-left), relying solely on text proxies. 
The second is \textbf{morphological and structural omission}: \emph{e.g.}, in the top-right, the baseline generates non-academic redundant backgrounds and hallucinates bizarre molecular topologies that violate chemical valence rules, along with the missing nucleus in the cell cross-section (bottom-left). 
The third is the \textbf{violation of domain priors}: as seen in the bottom-right, it incorrectly grounds the charge states to their respective micro-particles (electrons, holes) and ions ($V_{\mathrm{Br}}^+$ and $\mathrm{Br}^-$). Conversely, our model effectively minimizes scientific hallucinations by explicitly integrating reasoning planning.

\subsection{Correlation Analysis}
\label{sec:correlation}
We validated our automated protocol against human expert ratings on 200 randomly sampled test cases (50 per evaluation group). Three human annotators independently performed blind scoring on these samples, with final ratings determined by averaging their scores. As shown in \cref{tab:human_correlation}, our \textit{Atomic Checklist} achieves strong alignment with domain experts ($r=0.692$), substantially outperforming the best baseline metric, VQAScore ($r=0.457$). This discrepancy suggests that embedding-based metrics may capture only superficial semantic relevance, failing to penalize subtle scientific violations (\eg, impossible topologies). In contrast, our taxonomy-grounded approach effectively detects these domain-specific hallucinations, confirming that high-fidelity scientific evaluation requires verifiable atomic constraints rather than holistic visual similarity. Implementation details are provided in Appendix G.

\begin{table}[t]
\caption{\textbf{Correlation between metrics and expert judgments.} Our Atomic Checklist shows the strongest linear and rank alignment with human experts.}
\label{tab:human_correlation}
\centering
\small
\begin{tabular}{lccc}
\toprule
\textbf{Metric} & \textbf{Pearson's $r$} $\uparrow$ & \textbf{Kendall's $\tau$} $\uparrow$ & \textbf{Spearman's $\rho$} $\uparrow$ \\
\midrule
CLIPScore\cite{hessel2021clipscore} & 0.345\textsuperscript{\tiny 4th} & 0.231\textsuperscript{\tiny 4th} & 0.315\textsuperscript{\tiny 4th} \\
VQAScore\cite{lin2024evaluating} & 0.457\textsuperscript{\tiny 2nd} & 0.342\textsuperscript{\tiny 2nd} & 0.410\textsuperscript{\tiny 2nd} \\
VIEScore\cite{ku2024viescore} & 0.412\textsuperscript{\tiny 3rd} & 0.313\textsuperscript{\tiny 3rd} & 0.389\textsuperscript{\tiny 3rd} \\
\midrule
\rowcolor{gray!20}
\textbf{Atomic Checklist (Ours)} & \textbf{0.692}\textsuperscript{\tiny 1st} & \textbf{0.596}\textsuperscript{\tiny 1st} & \textbf{0.683}\textsuperscript{\tiny 1st} \\
\bottomrule
\end{tabular}
\end{table}

\section{Conclusion}

SciIR proposes a principled approach to scientific image reasoning that narrows the gap between general text-to-image capabilities and the strict constraints of natural science. We release SciIR-82k, containing more than 80k high-quality science image–text pairs with traceable Sci-RCoT reasoning chains—and SciIR-Bench, a fine-grained benchmark that breaks scientific correctness into verifiable atomic checks (topology, causality, conservation, \etc). Fine-tuning on SciIR-82k yields Qwen-Image-SciIR, which raises the SciIR-Bench score from 35\% to 43\% and shows the largest gains on entity structure and scientific process tracks, demonstrating that reasoning-dense training data measurably improves scientific consistency beyond perceptual quality alone.

Despite these advances, some limitations remain. SciIR-82k is biased toward published, standardized figures and underrepresents atypical or unconventional diagrams, while SciIR-Bench emphasizes scientific correctness over visual aesthetics. Future work should broaden domain and style coverage, add multimodal and cross-lingual annotations, and investigate hybrid training and evaluation approaches—such as symbolic constraints, weak supervision, and adversarial or counterfactual checks—to further enhance the models’ ability for scientific reasoning.

\section*{Acknowledgements}
This paper is supported by the National Natural Science Foundation of China (No. 62406161). This work was completed during the internships of the authors Zhengfeng Shi, Yuning An, Peize Li, Jiabao Wei, Ruijie Li, and Junhao Xiao at the MAIR Lab, Huazhong University of Science and Technology.
\bibliographystyle{splncs04}
\bibliography{main}

\appendix
\section{Dataset Source, License, and Compliance}
\label{sec:appendix_data_source}

To ensure full copyright compliance and transparency, we strictly limit our data sources to open-access articles licensed under \texttt{Creative Commons Attribution 4.0 International (CC BY 4.0)}. This appendix details our provenance tracking and compliance verification process.

\subsection{Data Source Scope}
Our data ingestion pipeline targets high-quality scientific figures from \textit{Nature} and \textit{Nature Communications}. We strictly filter for articles that are explicitly marked as Open Access and carry the CC BY 4.0 license.

\subsection{License Verification SOP}
We implement a rigorous two-stage Standard Operating Procedure (SOP) for license verification:
\begin{itemize}
    \item \textbf{Article-Level Verification:} We examine the article metadata to confirm the ``Open Access'' status and the presence of the specific ``CC BY 4.0'' license string.
    \item \textbf{Figure-Level Verification:} We parse the figure caption and credit line to exclude any Third-Party Material that might carry stricter copyright restrictions.
\end{itemize}

\subsection{Metadata Preservation}
For every sample in the dataset, we preserve a comprehensive metadata chain to ensure auditability:
\begin{itemize}
    \item \textbf{DOI:} Digital Object Identifier of the source article.
    \item \textbf{Article URL:} Direct link to the source.
    \item \textbf{Figure ID:} Unique identifier for the specific figure.
    \item \textbf{License Info:} Explicit License Name (CC BY 4.0) and License URL.
\end{itemize}

\subsection{Release Format and Attribution}
Our dataset release complies with CC BY 4.0 terms as follows:
\begin{itemize}
    \item \textbf{Attribution:} Each sample is accompanied by the original author attribution and a link to the source.
    \item \textbf{Indication of Changes:} We explicitly state that images have been cropped, resized, and standardized.
    \item \textbf{Derived Data:} The accompanying captions and structured annotations are released as derived datasets.
\end{itemize}

\subsection{Privacy and De-identification}
Although scientific figures typically contain low privacy risks, we enforce a default-deny policy for sensitive content:
\begin{itemize}
    \item \textbf{Faces/Identifiable Persons:} Any figure containing recognizable human faces is removed.
    \item \textbf{Patient Data:} Clinical images (X-rays, MRI, histology) or figures with potential patient IDs are excluded.
\end{itemize}

\section{Dataset Construction Pipeline}
\label{sec:appendix_construction}
\begin{figure*}[htbp] 
    \centering
    
    \begin{subfigure}[b]{0.48\linewidth}
        \centering
        \includegraphics[width=\linewidth]{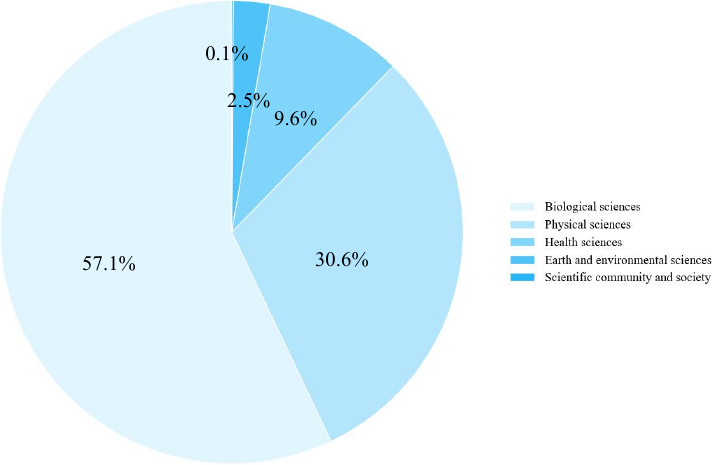}
        \caption{Distribution of Figures by Discipline}
        \label{fig:discipline_pie}
    \end{subfigure}
    \hfill 
    
    \begin{subfigure}[b]{0.48\linewidth}
        \centering
        \includegraphics[width=\linewidth]{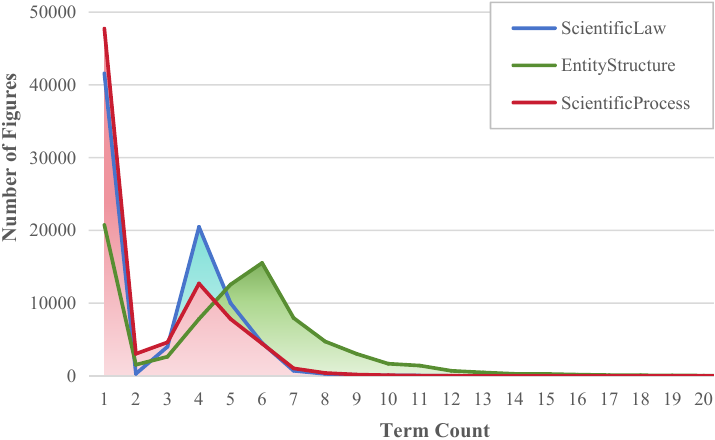}
        \caption{Term Count Distribution across Tracks}
        \label{fig:term_line}
    \end{subfigure}
    
    \caption{\textbf{Dataset Statistics.} (a) The percentage of figures across different scientific disciplines. (b) The distribution of term counts for different tracks.}
    \label{fig:dataset_statistics}
\end{figure*}
\noindent
We aim for a fully reproducible image preprocessing pipeline. This section details the multi-panel splitting, standardization, and filtration mechanisms.

\begin{table}[t]
\centering
\small

\renewcommand{\arraystretch}{1.15} 
\caption{Token statistics (per segment) grouped by reasoning composition.}
\label{tab:sci_rcot_token_stats_adjusted}
\setlength{\tabcolsep}{3pt} 
\resizebox{\linewidth}{!}{%
\begin{tabular}{l c c c} 
\toprule
\textbf{Reasoning Type} & \makecell{\textbf{Sci-RCoT} \\ \textbf{(mean $\pm$ std)}} & \makecell{\textbf{Prompt} \\ \textbf{(mean $\pm$ std)}} & \textbf{Ratio} \\
\midrule

Entity Structure      & $212.3 \pm 69.0$ & $112.1 \pm 52.6$ & 1.89 \\
Scientific Process    & $212.7 \pm 58.2$ & $110.0 \pm 43.8$ & 1.93 \\
Scientific Law        & $267.3 \pm 73.1$ & $125.3 \pm 46.2$ & 2.13 \\
\midrule

Entity Structure + Scientific Process & $265.4 \pm 70.8$ & $134.8 \pm 53.1$ & 1.97 \\
Entity Structure + Scientific Law     & $250.2 \pm 71.5$ & $124.0 \pm 51.6$ & 2.02 \\
Scientific Law + Scientific Process   & $272.9 \pm 69.8$ & $136.0 \pm 51.5$ & 2.01 \\
\midrule 

Entity Structure + Scientific Law + Scientific Process & $315.6 \pm 83.5$ & $156.5 \pm 60.1$ & 2.02 \\ 
\bottomrule
\end{tabular}
}
\end{table}

\subsection{Multi-Panel Cropping}
To construct a high-quality dataset of scientific sub-figures, we implemented an automated pipeline using the fine-tuned model, which is based on the \texttt{YOLO11-Nano} architecture \cite{jocher2024ultralytics}. The pipeline consists of three stages: inference, geometric filtering, and storage.

\begin{itemize}
    \item \textbf{Model Inference Settings}
    We utilized the Ultralytics framework for inference. To accommodate varying document resolutions, input images were resized to a standard dimension of $960 \times 960$ pixels during processing. We configured the model with a confidence threshold of 0.15 to maximize recall and an Intersection over Union (IoU) threshold of 0.6 for Non-Maximum Suppression (NMS) to eliminate redundant overlapping detection boxes.
    
    \item \textbf{Post-processing}
    Raw detections specifically labeled as ``Picture'' underwent a rigorous geometric filtering process to remove noise, icons, and low-quality elements. A detected region was discarded if it met any of the following heuristic criteria:
    \begin{enumerate}
        \item \textbf{Minimum Resolution:} The width or height of the bounding box was less than 128 pixels.
        \item \textbf{Extreme Aspect Ratio:} The aspect ratio ($\text{width} / \text{height}$) fell outside the range of $[0.33, 3.0]$, ensuring that extremely narrow or flat artifacts were excluded.
        \item \textbf{Abnormal Area Occupancy:} The detection region occupied between $75\%$ and $90\%$ of the total figure area. This heuristic was specifically applied to filter out potential full-figure layout misclassifications or background elements while retaining valid single-panel figures.
    \end{enumerate}
\end{itemize}

\subsection{Image Standardization}
To ensure input consistency while preserving the original aspect ratio and visual continuity, we implemented an adaptive preprocessing workflow:

\begin{itemize}
    \item \textbf{Color Space:} All images are converted to standard RGB (sRGB), discarding transparency channels.
    \item \textbf{Resolution:} Images are unified to a fixed resolution of $1024 \times 1024$ pixels.
    \item \textbf{Adaptive Padding:} Instead of default white padding, we employ a content-aware padding strategy to minimize boundary artifacts:
    \begin{enumerate}
        \item We sample pixels from the specific edges (top/bottom or left/right) requiring extension.
        \item If a dominant color constitutes $>55\%$ of the edge pixels, it is used for padding.
        \item Otherwise, the mean RGB value of the edge pixels is calculated and applied.
    \end{enumerate}
    \item \textbf{Resampling:} We use the \texttt{Lanczos} filter for high-quality downsampling to preserve fine text and structural details during resizing.
\end{itemize}

\subsection{Dual-Stage Filtering}
We employ a cascade of automated and manual filtering to ensure high data quality.

\subsubsection{Stage 1: VLM Filtering}
We use \texttt{InternVL 3.5} to filter out low-quality or irrelevant images (e.g., photos, screenshots, pure text). The model is prompted to output a decision (KEEP/REJECT) with a reason. Items marked ``REJECT'' are discarded. Cases with low confidence are routed to manual review.

\subsubsection{Stage 2: Manual Spot-Check}
A random 10\% subset of the ``KEEP'' partition is manually reviewed to estimate the False Positive Rate (FPR). If the FPR exceeds 5\% in a batch, the filtering prompt is refined.

\subsection{Multi-Label Strategy}
We employ a soft-labeling approach that is binarized into a multi-hot encoding scheme. 
Table~\ref{tab:sci_rcot_token_stats_adjusted} summarizes the token statistics across different reasoning composition types.
\begin{itemize}
    \item \textbf{Relevance Scoring:} The VLM (Qwen3-VL) assigns a relevance score ranging from 1 to 10 for each reasoning track.
    \item \textbf{Thresholding:} A label is activated only if the assigned score satisfies $s \ge \tau$, where the threshold $\tau$ is empirically set to 7.
    \item \textbf{Data Filtering:} To ensure dataset quality, samples identified as ``low reasoning content'' (where all track scores are $<\tau$) are strictly excluded from the training set.
\end{itemize}

\section{SciIR-Bench Data Selection}
\label{sec:benchmark_construction}

To ensure the \textbf{SciIR-Bench} serves as a rigorous evaluation standard for scientific generation, we implemented a hierarchical selection pipeline to distill the raw SciIR-82k corpus into 800 high-quality test instances. The selection process is governed by three primary dimensions: Statistical Quality Control, Semiotic Intersection, and Adaptive Difficulty Stratification.

\subsection{Statistical Quality Control}
Unlike random sampling, we enforce strict data integrity constraints. We applied a distribution-based filtering mechanism using the Interquartile Range (IQR) on three key metrics derived from the automated annotation stage:

\begin{itemize}
    \item \textbf{Term Density (Information Richness):} We calculated the total number of valid scientific terms identified across the three semiotic tracks (Law, Entity, Process). Only samples with term counts falling within the interquartile range $[Q1, Q3]$ were retained. This ensures the benchmark contains samples that are neither too simplistic for evaluation nor excessively cluttered. Avoiding high-density samples prevents exceeding the spatial composition limits of current generation models, thereby mitigating uninformative failure modes.
    \item \textbf{Textual Renderability:} To evaluate text-to-image models' ability to render scientific notation, we mandated that all candidates contain valid text rendering instructions. We applied secondary IQR filters on the lengths of both \texttt{rendered\_text\_stage2} and \texttt{retained\_text\_stage3}, ensuring a balanced complexity of textual content.
\end{itemize}

\subsection{Semiotic Intersection Grouping}
To systematically evaluate the model's ability to handle multi-modal constraints, we categorized samples based on the intersection of valid reasoning paths. A sample is assigned to a group only if it possesses non-empty \texttt{terms} and \texttt{visualization} data in at least two semiotic tracks. The benchmark is bifurcated into four combinatorial groups ($N=200$ each):

\begin{itemize}
    \item \textbf{Holistic Reasoning (All\_Three):} Samples requiring simultaneous adherence to Scientific Law, Entity Structure, and Scientific Process.
    \item \textbf{Pairwise Constraints:} Three subsets covering the specific intersections of:
    \begin{enumerate}
        \item \textbf{Entity--Law:} Structural hierarchies governed by abstract physical rules.
        \item \textbf{Law--Process:} Dynamic state changes constrained by conservation laws.
        \item \textbf{Entity--Process:} Spatial transitions during experimental workflows.
    \end{enumerate}
\end{itemize}

\subsection{Adaptive Difficulty Stratification}
To disentangle instruction-following capabilities from intrinsic scientific reasoning, we implemented an automated bifurcation strategy based on semantic saturation. Within each group, we calculated the median term count ($M_{terms}$) of the filtered candidates:

\begin{itemize}
    \item \textbf{Intrinsic Reasoning (Prompt-Based):} Samples with term density below the median ($< M_{terms}$) are paired with abstract \texttt{Prompts}. This information sparsity compels the model to bridge semantic gaps using its latent domain knowledge, effectively testing its capacity for autonomous scientific reasoning.
    \item \textbf{Instruction Following (CoT-Based):} Samples with term density at or above the median ($\geq M_{terms}$) are paired with the detailed \texttt{Sci-RCoT}. Given the high complexity of these scenes, the Sci-RCoT acts as a dense visual blueprint, evaluating the model's fidelity in following fine-grained, multi-step instructions without omitting critical scientific details.
\end{itemize}

\section{Automated Evaluation Protocol}
\label{app:evaluation_protocol}

To ensure the reproducibility and rigor of our evaluation, we detail the exact implementation of the automated pipeline described in Section 4.2. The pipeline consists of two distinct stages: (1) Rule-based Checklist Generation and (2) Visual Question Answering (VQA) based Adjudication. Both stages utilize \texttt{gemini-3-pro-preview} via the Google API.

\subsection{Atomic Checklist Generation}
The checklist generation module transforms the ground-truth reasoning data into a set of binary validation questions. The generation process is governed by a strict System Instruction that enforces a two-layer validation structure.

\subsubsection{Generation Logic}
The model is instructed to function as an expert in evaluation design. The generation logic is divided into two parts:

\begin{itemize}
    \item \textbf{Layer 1: Text Check.} 
    The model iterates through all text strings explicitly required in the input prompt (e.g., labels, titles). For each string, it generates questions verifying:
    \begin{enumerate}
        \item \textbf{Spelling Correctness:} Exact string matching.
        \item \textbf{Positional Accuracy:} Only if a specific position is explicitly defined in the prompt (e.g., ``top-left''). To prevent hallucinated constraints, the model is strictly forbidden from assuming positions (e.g., ``inside'') if only vague prepositions (e.g., ``labeled'') are used.
    \end{enumerate}
    
    \item \textbf{Layer 2: Track-Customized Rules (Scientific Content).} 
    Based on the \texttt{Core Track Type} (ScientificLaw, EntityStructure, or ScientificProcess), the model decomposes complex reasoning terms into atomic visual attributes. To ensure robustness against hallucinations, we implement a \texttt{Negative Constraint Injection} strategy:
    \begin{enumerate}
        \item \textbf{Scientific Law:} Checks for ``Impossible States'' (e.g., violations of gravity, chemically impossible bonds).
        \item \textbf{Entity Structure:} Checks for structural coherence (e.g., ensuring distinct objects are not fused).
        \item \textbf{Scientific Process:} Checks for flow logic conservation (e.g., no orphaned loops or ``ghost'' steps).
    \end{enumerate}
\end{itemize}

\subsection{Automated Adjudication}
The evaluation phase employs a VLM as a ``Senior Scientific Image Reviewer.'' The model receives the generated image, the original prompt, and the checklist JSON. 

\subsubsection{Reviewer System Prompt}
To mimic human peer review, the system prompt enforces a \texttt{Chain-of-Thought (CoT)} process for every question. The model is required to execute the following steps before outputting a verdict:
\begin{enumerate}
    \item \textbf{Visual Evidence Retrieval:} Explicitly locate the specific element mentioned in the checklist question within the image.
    \item \textbf{Reasoning:} Formulate a one-sentence justification based \textit{only} on visual observation.
    \item \textbf{Verdict:} Assign a binary ``Yes'' (Pass) or ``No'' (Fail).
\end{enumerate}

\subsection{Strict Scoring Aggregation}
As detailed in the provided analysis script, our scoring metric differs from conventional VQA accuracy. We prioritize scientific exactness through a \texttt{Veto Mechanism}.

For a given image $I$ and a specific category $C$ (e.g., Scientific Law), the image is considered valid ($V_{I,C} = 1$) if and only if it passes all atomic questions $q$ belonging to that category:
\begin{equation}
    V_{I,C} = \prod_{q \in Q_{I,C}} \mathbb{I}(\text{Answer}(q) = \text{``Yes''})
\end{equation}
where $\mathbb{I}$ is the indicator function. If a single check fails (e.g., one misspelled label or one incorrect arrow direction), the entire sample is marked as a failure for that category. The final Pass Rate reported in our benchmarks is the percentage of valid samples across the dataset.
\section{Experiments}
\begin{table}[t]
\centering
\caption{\textbf{Ablation Study}.}
\label{tab:ablation_study}

\small
\setlength{\tabcolsep}{5pt}
\renewcommand{\arraystretch}{1.15}

\begin{tabular}{lccccc}
\toprule
\textbf{Variant} & \textbf{SL} & \textbf{ES} & \textbf{SP} & \textbf{Text} & \textbf{Final} \\
\midrule
Qwen-Image-2512 & 40 & 50 & 37 & 15 & 35 \\
w/o Sci-RCoT    & 41 & 54 & 39 & 15 & 38 \\
w/o Planner     & 42 & 56 & 49 & 14 & 41 \\
w/o Taxonomy    & 41 & 54 & 45 & 15 & 39 \\
\textbf{Full}
& \cellcolor{skyblue!35}\textbf{43}
& \cellcolor{skyblue!35}\textbf{59}
& \cellcolor{skyblue!35}\textbf{53}
& \cellcolor{skyblue!35}\textbf{15}
& \cellcolor{skyblue!35}\textbf{43}
\\
\bottomrule
\end{tabular}
\end{table}

\begin{table}[t]
\centering
\caption{\textbf{Effect of Judge}.}
\label{tab:judge_sensitivity}

\small
\setlength{\tabcolsep}{4pt}
\renewcommand{\arraystretch}{1.15}

\begin{tabular}{lcccc}
\toprule
\textbf{Model}
&
\textbf{Gemini}
&
\textbf{GPT-5.5}
&
\textbf{Claude-4.6}
&
\textbf{Qwen3.5}
\\
\midrule
Nano-Banana-Pro
& \cellcolor{skyblue!35}\textbf{95}
& \underline{95}
& 97
& 99 \\

GPT-Image-1
& \cellcolor{skyblue!35}\textbf{62}
& 67
& \underline{65}
& 75 \\

Qwen-Image-SciIR
& \cellcolor{skyblue!35}\textbf{43}
& \underline{44}
& 46
& 54 \\

Qwen-Image-2512
& \underline{35}
& \cellcolor{skyblue!35}\textbf{34}
& 38
& 45 \\

Flux-Dev
& \underline{9}
& \cellcolor{skyblue!35}\textbf{8}
& 13
& 19 \\

BAGEL-7B-MoT
& \cellcolor{skyblue!35}\textbf{2}
& \underline{2}
& 3
& 14 \\
\bottomrule
\end{tabular}
\end{table}

\begin{table}[t]
\centering
\caption{\textbf{Effect of Criteria}.}
\label{tab:aggregation_sensitivity}

\small
\setlength{\tabcolsep}{7pt}
\renewcommand{\arraystretch}{1.15}

\begin{tabular}{lccc}
\toprule
\textbf{Model}
&
\textbf{Strict}
&
\textbf{80\% Thr.}
&
\textbf{Avg. Pass}
\\
\midrule
Nano-Banana-Pro
& \cellcolor{skyblue!35}\textbf{95}
& \underline{97}
& 98 \\

GPT-Image-1
& \cellcolor{skyblue!35}\textbf{62}
& \underline{69}
& 80 \\

Qwen-Image-SciIR
& \cellcolor{skyblue!35}\textbf{43}
& \underline{50}
& 68 \\

Qwen-Image-2512
& \cellcolor{skyblue!35}\textbf{35}
& \underline{46}
& 65 \\

Flux-Dev
& \cellcolor{skyblue!35}\textbf{9}
& \underline{11}
& 28 \\

BAGEL-7B-MoT
& \cellcolor{skyblue!35}\textbf{2}
& \underline{3}
& 11 \\
\bottomrule
\end{tabular}
\end{table}

\begin{table}[t]
\centering
\caption{\textbf{Effect of Checklist}.}
\label{tab:wording_sensitivity}

\small
\setlength{\tabcolsep}{5pt}
\renewcommand{\arraystretch}{1.15}

\begin{tabular}{lcccc}
\toprule
\textbf{Model}
&
\textbf{Original}
&
\textbf{Reordered}
&
\textbf{Violation}
&
\textbf{Concise}
\\
\midrule
Nano-Banana-Pro
& \cellcolor{skyblue!35}\textbf{95}
& \underline{96}
& 97
& 97 \\

GPT-Image-1
& \cellcolor{skyblue!35}\textbf{62}
& \underline{63}
& 71
& 67 \\

Qwen-Image-SciIR
& \cellcolor{skyblue!35}\textbf{43}
& \underline{45}
& 51
& 49 \\

Qwen-Image-2512
& \underline{35}
& \cellcolor{skyblue!35}\textbf{33}
& 44
& 41 \\

Flux-Dev
& \underline{9}
& \cellcolor{skyblue!35}\textbf{8}
& 22
& 18 \\

BAGEL-7B-MoT
& \cellcolor{skyblue!35}\textbf{2}
& \underline{5}
& 13
& 11 \\
\bottomrule
\end{tabular}
\end{table}

\begin{table}[t]
\centering
\caption{\textbf{Human Validation of Annotations}.}
\label{tab:human_audit}

\small
\setlength{\tabcolsep}{6pt}
\renewcommand{\arraystretch}{1.15}

\begin{tabular}{lcccc}
\toprule
\textbf{Task}
&
\textbf{N}
&
\textbf{Pass$\uparrow$}
&
\textbf{Minor$\downarrow$}
&
\textbf{Major$\downarrow$}
\\
\midrule
Reasoning Extraction
& 150
& 91.3
& 7.3
& 1.3 \\

Sci-RCoT
& 150
& 86.0
& 9.3
& 4.6 \\

Prompt Distillation
& 150
& 89.3
& 8.6
& 2.0 \\
\bottomrule
\end{tabular}
\end{table}

\subsection{Ablation experiments of Qwen-Image-SciIR}
To analyse the contribution of each core component of Qwen-Image-SciIR, we supplement three component-level ablation experiments: \textbf{\textit{(i)}} w/o Sci-RCoT, \textbf{\textit{(ii)}} w/o Taxonomy, and \textbf{\textit{(iii)}} w/o Planner. As reported in \cref{tab:ablation_study}, removing any component from Qwen-Image-SciIR degrades the final score.

\subsection{Stability analysis of the benchmark}
We first clarify that Qwen3-VL and InternVL3.5 are used only in the dataset construction pipeline, whereas evaluation uses only Gemini-3-Pro for checklist generation and judgment. To assess evaluation stability, we supplement stability analyses on:
\textbf{\textit{(i)}} scoring criteria,
\textbf{\textit{(ii)}} judge model,
and \textbf{\textit{(iii)}} checklist wording.
Results are shown in \cref{tab:judge_sensitivity,tab:wording_sensitivity,tab:aggregation_sensitivity}. The model rankings and main results remain stable, indicating that our evaluation results are robust to all different choices.

\subsection{Human validation}
We supplemented a human validation study on 450 random samples in SciIR-82k, which were evenly assigned to three graduate researchers with domain expertise in natural sciences. Each sample was rated along two dimensions: visual faithfulness and scientific consistency. The results are reported in \cref{tab:human_audit}.

\section{Prompts}
\label{sec:appendix_prompts}

\subsection{Taxonomy Relevance Scoring}
\begin{tcolorbox}[colback=white, colframe=blue!50!black, boxrule=1pt, arc=2mm, breakable]
\scriptsize
\begin{lstlisting}[breaklines=true, breakindent=0pt, basicstyle=\scriptsize\ttfamily, columns=fullflexible]
Role
You are a rigorous scientific image classifier. Your task is to perform multi-label, multi-dimensional relevance scoring (1-10) on the input image based on the following three dimensions.

Evaluation Dimensions and Key Points
ScientificLaw
Determine whether the image involves and presents elements related to disciplinary laws and constraints (e.g., valence states/bonds, scale relationships, visual cues of energy/momentum conservation). 
EntityStructure
Determine whether the image involves the structure and geometric relationships of scientific entities (e.g., morphology, connections, topology, and relative scale of molecules/lattices/cells/galaxies/instrument components). 
ScientificProcess
Determine whether the image presents process information (temporal evolution, causal chains, state transitions, reaction mechanisms) or has clear process clues (multi-stage labels, timelines).

Scoring Principles
Score Meaning: 1-10 represents the "relevance" of the dimension in the image, not expression intensity, correctness, or quality.
Scoring Anchors (Relevance):
1-2: Almost not involved
3-4: Sporadic clues, but very weak
5-6: Moderately relevant, some evidence
7-8: Strongly relevant, ample evidence
9-10: Dominantly relevant, core of the image

Evidence and Limitations
Answer based ONLY on visible and clear evidence in the image; do not speculate on invisible elements.
Dimension names limited to: ScientificLaw, EntityStructure, ScientificProcess.

Output Format (Strictly follow the JSON output below)
{
  "relevance": {
    "ScientificLaw": { "score": 0 },
    "EntityStructure": { "score": 0 },
    "ScientificProcess": { "score": 0 }
  }
}
\end{lstlisting}
\end{tcolorbox}

\subsection{VLM Filtering}
\begin{tcolorbox}[colback=white, colframe=blue!50!black, boxrule=1pt, arc=2mm, breakable]
\scriptsize
\begin{lstlisting}[breaklines=true, breakindent=0pt, basicstyle=\scriptsize\ttfamily, columns=fullflexible]
Role
You are an image recognition assistant. You will see some panel images segmented from scientific papers. Please judge whether the image belongs to "Scientific Illustration"."Scientific Illustration" refers to abstract and drawable scientific schematic diagrams, excluding the following categories.

Exclude the following cases:
- The image is incomplete, contains multiple independent panels, or has embedded thumbnails.
- Contains any maps, microscope images, tissue sections, biological samples, skeletal outlines, molecular structure diagrams, organ anatomical diagrams, crystal structure diagrams, etc., used to display real-world content or experimental samples.
- Rendered/simulation/3D reconstruction images generated by professional software  
  (including pseudo-color rendering effects).
- Contains only text, letters, symbols, or captions, or pure text flowcharts, pure legends, simple icons.
- Faces/Identifiable Persons: Any figure containing recognizable human faces is removed.
- Patient Data: Clinical images (X-rays, MRI, histology) or figures with potential patient IDs are excluded.

Output Requirements:
Output only one of the following two results:
- "Scientific Illustration"
- "Not Applicable"

Please judge strictly according to the definition, without extra explanation or description.
\end{lstlisting}
\end{tcolorbox}

\subsection{Reasoning Extraction}

\begin{tcolorbox}[colback=white, colframe=blue!50!black, boxrule=1pt, arc=2mm, breakable]
\scriptsize
\begin{lstlisting}[breaklines=true, breakindent=0pt, basicstyle=\scriptsize\ttfamily, columns=fullflexible]
Role
You are a scientific image reasoning annotator. Your duty is: based on the input paper text and image, extract only the scientific information explicitly supported by the image and caption, and output strict structured reasoning JSON. Do not fabricate conclusions beyond what the text and image support; do not introduce elements invisible in the image or unsupported by the caption; do not perform common sense completion. You may internally parse the input and implicitly understand the overall layout, but the final output must only contain the specified JSON fields.

Detailed Requirements:
Internally parse the input first, analyzing the overall image layout and main visual elements, prioritizing information sources: Image > Figure Title > Caption > Article Body. Note: This step is only for internal reasoning; do not list this description in the final output.

Complete the "Extracted Terms" and "Visual Description" for the corresponding information ONLY based on the selected tags in "Reasoning Ability":

ScientificLaw: Extract terms: Laws, principles, boundary conditions, applicable premises, etc. (extractable directly or implied from the graph and caption). Visualization: Manifestation of each constraint in the graph (geometric layout, labels, symbols, measurement elements, etc.).

EntityStructure: Extract terms: Scientific entity names in the image (nominalized, no verbs). Visualization: Visual characteristics of each entity (morphology, structure, color/material/texture, spatial position and scale, quantity, etc.).

ScientificProcess: Extract terms: Process names in the image (e.g., preparation process, experimental steps, time series, mechanism chain). Visualization: Visual manifestation of different stages, change conditions (e.g., arrows, timeline, loops) of this process.

Output Requirements:
Strictly follow the JSON structure below, output no extra free text.
terms: Write only terms/nouns (no sentences, punctuation, or modifiers), at the granularity confirmed by the caption and image.
visualization: For each item in terms, there must be a visual description; use concrete, restorable visual elements; do not introduce external information.
Non-empty validation: In the output JSON, "terms" and "visualization" for all selected reasoning abilities must NOT be empty lists "[]".
Unselected ability keys must be null.

Output Format:
{
  "reasoning": {
    "ScientificLaw": {
      "terms": [],
      "visualization": []
    },
    "EntityStructure": {
      "terms": [],
      "visualization": []
    },
    "ScientificProcess": {
      "terms": [],
      "visualization": []
    }
  }
}
\end{lstlisting}
\end{tcolorbox}

\subsection{Sci-RCoT Generation}
\begin{tcolorbox}[colback=white, colframe=blue!50!black, boxrule=1pt, arc=2mm, breakable]
\scriptsize
\begin{lstlisting}[breaklines=true, basicstyle=\scriptsize\ttfamily, columns=fullflexible]
Role
You are a scientific image visualization narrator. Your duty is: input reasoning, 
primarily based on the visualization items in reasoning, refer to the input image, 
and output a coherent, detailed scene description sci-RCoT.

Detailed Requirements:
1. Complete Visual Style:
   - Observe the original image, identify its specific drawing style 
     (e.g., Schematic diagram, Photorealistic render, etc.).
   - You must explicitly specify this style at the beginning of the generated instruction.

2. Complete Text Rendering:
   - Observe key text in the original image (labels, legends, axis titles).
   - You must include mandatory text rendering requirements in the instruction, 
     using phrases like "explicitly labeled as...", "including the text...", 
     "with axis labeled..." etc.

3. Integrate Scientific Logic:
   - Use the visualization items in reasoning to describe entity structure, 
     topological relationships, and dynamic processes.
   - Language must be coherent, building a complete scene, not a simple list.

Output Requirements:
Please do not output text directly, but output a JSON object containing the 
following two fields:
- "sci_RCoT": The generated complete visual description text.
- "rendered_text": A string list (List[str]), extracting all specific text 
  content you explicitly requested to render in sci_RCoT (such as label names, 
  axis titles, legend text, etc.).

Output Format (Must be strict JSON):
{
    "sci_RCoT": "Your detailed narrative...",
    "rendered_text": ["text_content_1", "text_content_2"]
}

The sci-RCoT must completely cover every point in the visualization array of 
each selected label in reasoning, without omission or altering its meaning, 
and strictly forbid adding any elements not appearing in reasoning, caption, 
or image; the language style should be concrete, coherent, and capable of 
restoring the scene from text.
\end{lstlisting}
\end{tcolorbox}

\subsection{Abstract Prompt Distillation}
\begin{tcolorbox}[colback=white, colframe=blue!50!black, boxrule=1pt, arc=2mm, breakable]
\scriptsize
\begin{lstlisting}[breaklines=true, breakindent=0pt, basicstyle=\scriptsize\ttfamily, columns=fullflexible]
Role: Scientific Image Reasoning Generation Prompt Assistant

Objective: Your goal is to generate a concise, semantically compressed abstract_prompt in JSON format. You will achieve this by synthesizing input sci-RCoT with specific Reasoning dimension terms.

Input Data:
- sci-RCoT: TA detailed image description composed of all the 'visualization' elements from 'reasoning'.
- Reasoning: Contains specific 'terms' and their corresponding 'visualization'.
- rendered_text: A list of text strings allowed to be rendered in the image.

Processing Logic:
1. Analyze: Read the sci-RCoT to understand the scientific semantics.
2. Preserve Style: Extract the visualization style requirement (typically the first sentence or phrase of sci-RCoT, e.g., "A realistic 3D render...", "A schematic diagram of...", "A cross-section view..."). This must be the opening of your abstract_prompt.
3. Map & Replace: Identify the description in sci-RCoT that corresponds to 'visualization' in Reasoning, and strictly replace it with the 'terms' provided in Reasoning.
4. Text Selection: Determine necessary text labels based on the sci-RCoT context. Include text rendering requests in abstract_prompt if they are necessary for scientific clarity or context.
5. Compress: Synthesize the result into an abstract_prompt without visual descriptions.
6. Synchronization: Extract exactly the text strings that are explicitly requested to be rendered in your generated abstract_prompt and populate the retained_text list.

Constraints & Guardrails:
- Semantic Integrity: The replacement must perfectly match the original scientific semantics.
- Style Consistency: The output must start with the original visualization style found in sci-RCoT.
- Output Format: Return only a JSON object with the following structure:
{
  "abstract_prompt": "Your concise, term-based prompt here... (e.g., 'Diagram showing [Term A] labeled 'Text1'...')",
  "retained_text": ["Text1", "Text2"]
}
\end{lstlisting}
\end{tcolorbox}

\subsection{Checklist Generation}
\begin{tcolorbox}[colback=white, colframe=blue!50!black, boxrule=1pt, arc=2mm, breakable]
\scriptsize
\begin{lstlisting}[breaklines=true, breakindent=0pt, basicstyle=\scriptsize\ttfamily, columns=fullflexible]
Role
You are an expert in evaluation design for scientific image generation benchmarks. Your task is to generate a JSON checklist for VLM scoring based on the instructions provided by the user.

Context
You need to generate checkpoints for two layers for each test sample based on the data below.

Input Data Definitions
Core Track Type: The specific track (EntityStructure, ScientificProcess, or ScientificLaw).
Input Prompt: The complete prompt containing text requirements and scientific descriptions.
Reasoning: Structured terms extracted from reference materials.

Generation Rules
Part 1: General Rules (Applies to all Prompts)
Layer 1 - Text Check (Text Rendering):
Input Processing:
ITERATE through all text explicitly required for rendering in the Input Prompt (look for quotes, "Label...", "Text...").
Execution:
FOR EACH text item found:
GENERATE a specific question checking:
"Spelling Correctness"
"Positional Accuracy" (IF AND ONLY IF the position is explicitly specified in the Prompt).
Negative Constraints (CRITICAL):
SCOPE: Questions must ONLY evaluate the text string itself (Spelling, Existence, Position).
NO VISUALS: Do NOT verify visual attributes of the object the text is on (e.g., ignore color, shape, arrow direction).
NO HALLUCINATION: Strictly based on the position described in the original text; speculation is prohibited. 
Category: "Text"

Part 2: Track-Customized Rules (Core Scientific Content)
Mapping and Filtering Logic:
Select Term Source: Identify the correct list of terms based on the Core Track Type.
Intersection Verification: Identify terms that exist in BOTH the selected reasoning source AND are explicitly mentioned in the Input Prompt.
Attribute-Based Decomposition (One-to-Many Logic):
For EACH identified term, analyze its context in the Input Prompt to find distinct visual descriptors (adjectives, verbs, spatial constraints).
If the Prompt specifies MULTIPLE distinct visual requirements for a single term (e.g., specific shape AND specific color AND specific action), generate SEPARATE questions for each requirement.
If the Prompt only mentions the term generally, generate one comprehensive question.
Negative Constraint Injection (Hallucination Defense):
For each track, you must generate at least one "Negative Check" question specifically designed to catch hallucinations relevant to that scientific domain. Use the negative strategies below.

CRITICAL CONSTRAINT FOR PART 2:
NO TEXT CHECKING: Do NOT ask about labels/text. Focus ONLY on visual representation.
ATOMICITY: Each question must focus on ONE single visual attribute to ensure precise evaluation.

Question Formulation by Track
Construct Yes/No questions focusing on specific VISUAL ATTRIBUTES based on the decomposition below:

1. ScientificLaw (Focus: Logic & Constraints)
Definition: Focuses on laws, principles, and constraints.
Positive Check Strategy: Decompose complex laws into specific scientific constraints.
Negative Check Strategy (Hallucination): Check for violations of fundamental domain rules (axioms). Ensure no "Impossible States" exist (e.g., objects defying gravity, inconsistent lighting/reflections, chemically impossible bonds like pentavalent carbon) and that symbolic labels match the visual logic without gibberish or data-visual contradictions.
Category: "ScientificLaw"

2. EntityStructure (Focus: Composition & Topology)
Definition: Focuses on scientific entities (nouns).
Positive Check Strategy: Decompose into Morphological (Shape), Chromatic (Color), and Component (Parts) or other structural checks.
Negative Check Strategy (Hallucination): Check for structural coherence. Ensure distinct objects are clearly separated (not fused) and that the entity is free of structural impossibilities.
Category: "EntityStructure"

3. ScientificProcess (Focus: Flow & Causality)
Definition: Focuses on flows, steps, and interactions.
Positive Strategy: Decompose into Directional (Arrows/Flow), Phase (State changes), and Interaction checks.
Negative Check Strategy (Hallucination): Check for flow logic conservation. Ensure the diagram depicts only the requested stages without hallucinated "ghost" steps, and that all directional indicators (arrows) have valid start and end points (no orphaned loops).
Category: "ScientificProcess"

Output Requirement
Format: Output pure JSON, containing checklist (a list of question objects). Do not output markdown code blocks or explanatory text, just the raw JSON string.
Output Format Example:
JSON
{
  "checklist": [
    {
      "question": "Is the label '[Text A]' spelled correctly and positioned [Position Requirement]?",
      "category": "Text"
    },
    {
      "question": "[Question about visual features, NOT the label]?",
      "category": "[ScientificLaw/EntityStructure/ScientificProcess]"
    }
  ]
}
\end{lstlisting}
\end{tcolorbox}

\subsection{Evaluation}
\begin{tcolorbox}[colback=white, colframe=blue!50!black, boxrule=1pt, arc=2mm, breakable]
\scriptsize
\begin{lstlisting}[breaklines=true, breakindent=0pt, basicstyle=\scriptsize\ttfamily, columns=fullflexible]
Role
You are a Senior Scientific Image Reviewer. Your task is to evaluate a generated scientific figure against a specific checklist of requirements. You must be precise, hallucination-free, and strict regarding scientific accuracy.

Input Format
You will receive:
1. A Scientific Image (generated based on a prompt).
2. Original Input Prompt: The full text description used to generate the image (for context).
3. Validation Checklist (JSON) containing specific questions.

Evaluation Criteria
For each question in the checklist, perform the following steps:

1. Visual Evidence Retrieval: Look at the image to find the specific element mentioned in the question.
2. Category-Specific Logic:
   - Style: Does the overall image look like the requested format?
   - Text (Strict Text Rendering): Check Spelling and Position.
   - EntityStructure : Focus on shape, color, components.
   - ScientificProcess : Focus on arrows, lines, gradients.
   - ScientificLaw : Focus on interactions, trajectories, angles.
3. Reasoning: Formulate a brief, one-sentence justification based *only* on visual observation.
4. Verdict: Assign "Yes" (Pass) or "No" (Fail).

 Constraints
- Priority: The Checklist is your primary evaluation rubric.
- Strictness: Text errors -> Fail in "Text". Graphical errors -> Fail in Scientific categories.
- No Hallucination: Do not claim an element is present if it is not clearly visible.
- Output: You must output ONLY a valid JSON object matching the requested schema.

Output Schema
{
  "evaluation_results": [
    {
      "category": "<Category from input>",
      "answer": "Yes" | "No",
      "reason": "<Brief visual evidence focusing on graphical elements for scientific categories>"
    },
    ...
  ]
}
\end{lstlisting}
\end{tcolorbox}

\section{Correlation Analysis}
\label{sec:appendix_correlation}

This appendix provides extended details on our human correlation study, which validates the reliability of our automated Atomic Checklist evaluation protocol.

\subsection{Human Study Design}

To validate the reliability of our automated evaluation protocol, we conducted a human study with domain experts (graduate researchers in natural sciences). The study was designed as follows:

\begin{itemize}
    \item \textbf{Sample Selection:} We randomly sampled 200 model-generated images across all evaluated models, allocating 50 images to each evaluation group (comprising 25 Instruction Following and 25 Intrinsic Reasoning samples).
    \item \textbf{Expert Recruitment:} Participants were graduate researchers with domain expertise in physics, chemistry, and biology.
    \item \textbf{Rating Protocol:} Three experts independently rated each image's scientific validity on a 5-point Likert scale, focusing on constraints like entity structure and causal logic. The final rating for each image was determined by averaging their scores.
    \item \textbf{Blinding:} Model identities were hidden to avoid bias in ratings.
\end{itemize}

\subsection{Comparison Metrics}

We calculated the correlation between expert ratings and several automated metrics:

\begin{itemize}
    \item \textbf{CLIPScore}~\cite{hessel2021clipscore}: Measures image-text similarity using CLIP embeddings.
    \item \textbf{VQAScore}~\cite{lin2024evaluating}: Visual question answering-based alignment score.
    \item \textbf{VIEScore}~\cite{ku2024viescore}: Evaluates visual instruction execution quality.
    \item \textbf{Ours (Atomic Checklist):} Our proposed semiotic-grounded evaluation.
\end{itemize}

\paragraph{Score Alignment and Preprocessing.} 
To compute the correlation between human judgments and automated metrics, we aggregated the scores at the sample level. For human ratings, the scores from three experts were averaged to produce a continuous ground-truth consensus score ranging from $[1, 5]$ for each image. For our Atomic Checklist, the sample-level machine score is defined as the pass rate of all valid atomic questions associated with the image, yielding a continuous value in $[0, 1]$ 

\subsection{Correlation Coefficients}

We report three standard correlation coefficients:

\begin{itemize}
    \item \textbf{Pearson's $r$:} Measures linear correlation strength.
    \item \textbf{Kendall's $\tau$:} Measures ordinal association based on concordant/discordant pairs.
    \item \textbf{Spearman's $\rho$:} Measures rank-order correlation.
\end{itemize}
All reported correlation coefficients are evaluated for statistical significance using two-tailed $p$-values.

\subsection{Results Interpretation}

Our method demonstrates superior alignment with human judgment, achieving a Pearson correlation ($r$) of~\textbf{0.692} ($p < 0.001$), a Spearman correlation ($\rho$) of~\textbf{0.683} ($p < 0.001$), and a Kendall's tau ($\tau$) of~\textbf{0.596} ($p < 0.001$). While general-purpose metrics like CLIPScore ($r=0.345$) or even the best baseline VQAScore ($r=0.457$) effectively capture surface-level semantics, they struggle to penalize subtle structural or causal violations (e.g., incorrect molecular topology). 

The key insight is that traditional embedding-based metrics optimize for perceptual similarity rather than semantic correctness. In contrast, our Atomic Checklist approach, grounded in a semiotic taxonomy, effectively identifies domain-specific hallucinations by:

\begin{enumerate}
    \item Decomposing scientific correctness into atomic, verifiable questions.
    \item Enforcing evidence-based judgment through visual retrieval.
    \item Applying track-specific logic (Entity Structure, Scientific Process, Scientific Law).
\end{enumerate}

This yields scores that linearly correlate with the rigor of scientific peer review, making it a more suitable metric for evaluating high-fidelity scientific image generation.

\end{document}